\documentclass[11pt, a4paper, logo, copyright, nonumbering]{lti}
\usepackage[authoryear, sort&compress, round]{natbib}
\usepackage{dblfloatfix}
\usepackage{ulem}
\usepackage{caption}
\usepackage{dramatist}
\usepackage{xspace}
\usepackage{multirow}
\usepackage{tcolorbox}
\usepackage{xltabular}
\usepackage{longtable}
\usepackage{hyperref}
\usepackage{amsfonts}
\usepackage{amsmath}
\usepackage{amssymb}
\usepackage{adjustbox}
\usepackage[bottom]{footmisc}
\usepackage{CJKutf8}
\usepackage{setspace}
\usepackage{dsfont}
\usepackage{array}
\usepackage{tabularx}
\usepackage{xcolor}
\usepackage{wrapfig}
\usepackage{booktabs}

\definecolor{ltired}{HTML}{780000}
\usepackage{lipsum}
\usepackage{multicol}
\usepackage{makecell}
\usepackage{graphicx}
\usepackage{color, colortbl}

\usepackage{float}

\newcommand{\cp}{\textsc{CowPilot}\xspace}
\newcommand{\cc}{\textsc{CowCorpus}\xspace}
\newcommand{\cpnew}{\textsc{PlowPilot}\xspace}
\newcommand{\zora}[1]{\textcolor{orange}{Zora: #1}}

\usepackage{url}

\usepackage{pifont}
\usepackage{fontawesome5}
\usepackage{subcaption}
\usepackage[table]{xcolor}
\tcbuselibrary{skins}
\tcbset{tabtag/.style={
  enhanced,
  colback=#1!20,
  colframe=#1!60!black,
  boxrule=0.5pt,
  arc=2mm,
  boxsep=1pt,
  left=2pt,
  right=2pt,
  top=1pt,
  bottom=1pt,
  fontupper=\scriptsize
}}

\newtcolorbox{quoteBox}{
    colback=gray!10,
    colframe=gray!50,
    boxrule=0.5pt,
    arc=2mm,
    left=2mm,
    right=2mm,
    top=1mm,
    bottom=1mm,
    fontupper=\small\itshape,
    width=0.95\columnwidth,
    boxsep=1mm
}

\usepackage{xspace}

\usepackage{color-edits}
\addauthor{gn}{magenta}

\definecolor{group0}{HTML}{40a394}
\definecolor{group1}{HTML}{919a8f}
\definecolor{group2}{HTML}{eb9058}
\definecolor{group3}{HTML}{e87c7e}

\newcommand{\cmark}{\textcolor{green}{\ding{51}}}
\newcommand{\xmark}{\textcolor{red}{\ding{55}}}

\newcommand{\icon}[1]{\raisebox{-0.25\height}{\includegraphics[height=1em]{#1}}}

\newcommand{\cmu}{\mathbb{C}}
\newcommand{\duke}{\mathbb{D}}

\title{\centering Modeling Distinct Human Interaction in Web Agents}

\author{\centering
\normalsize{\textbf{Faria Huq$^{\cmu,\dagger}$ \quad Zora Zhiruo Wang$^{\cmu,\dagger}$ \quad Zhanqiu Guo$^{\cmu,\ddagger}$ \quad Venu Arvind Arangarajan$^{\cmu,\ddagger}$}} \\
\normalsize{\textbf{Tianyue Ou$^{\cmu}$ \quad Frank Xu$^{\cmu}$ \quad Shuyan Zhou$^{\duke}$ \quad Graham Neubig$^{\cmu}$ \quad Jeffrey P. Bigham$^{\cmu}$}} \\
\vspace{0.3em}
$^{\cmu}$Carnegie Mellon University \quad $^{\duke}$Duke University 
\\
\vspace{0.3em}
{\small{$^\dagger$ Co-first Authors\quad $^\ddagger$ These authors contributed equally}}\\
\small{\texttt{\{fhuq, zhiruow\}@cs.cmu.edu}}\\
\vspace{0.3em}
\vspace{0.7em}
\texttt{\raisebox{-1ex}{\includegraphics[width=20pt]{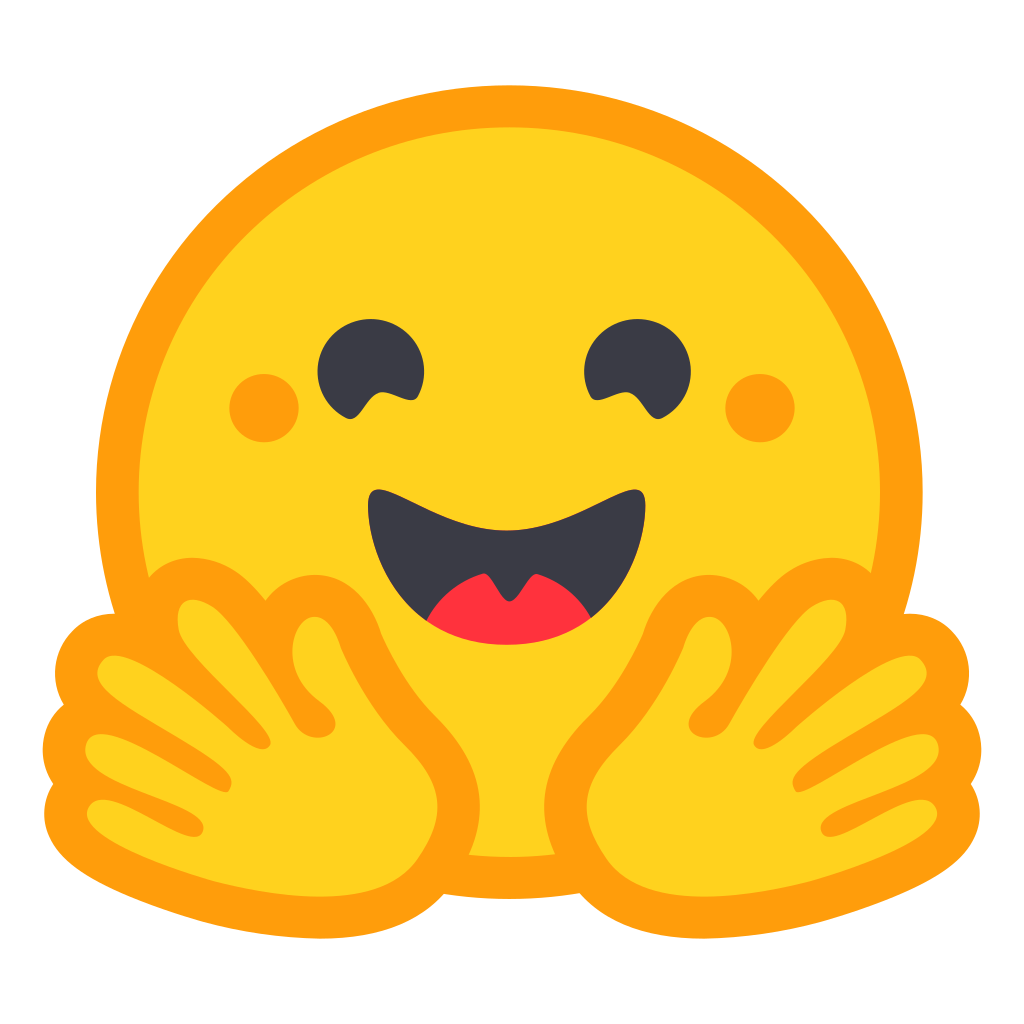}} Models:\href{https://huggingface.co/CowCorpus}{huggingface.co/CowCorpus}}
~\texttt{\raisebox{-1ex}{\includegraphics[width=20pt]{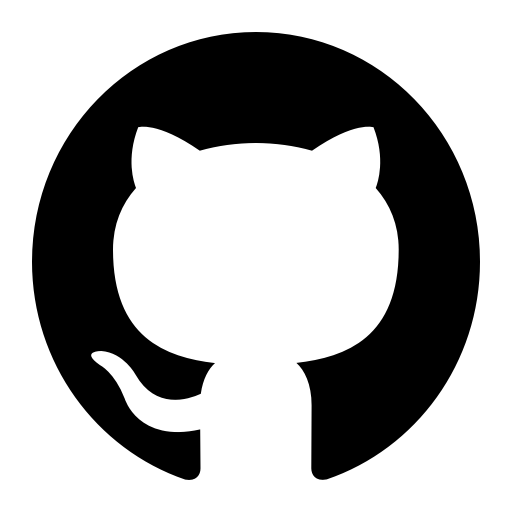}} Code:\href{https://github.com/oaishi/PlowPilot}{github.com/oaishi/PlowPilot}}
}

\begin{abstract}
Despite rapid progress in autonomous web agents, human involvement remains essential for shaping preferences and correcting agent behavior as tasks unfold. However, current agentic systems lack a principled understanding of \textit{when} and \textit{why} humans intervene, often proceeding autonomously past critical decision points or requesting unnecessary confirmation.
In this work, we introduce the task of \textit{modeling human intervention} to support collaborative web task execution. We collect \cc{}, a dataset of 400 real-user web navigation trajectories containing over 4{,}200 interleaved human and agent actions.
We identify four distinct patterns of user interaction with agents -- \textcolor{group1}{\textbf{hands-off}} supervision, \textcolor{group2}{\textbf{hands-on}} oversight, \textcolor{group0}{\textbf{collaborative}} task-solving, and full user \textcolor{group3}{\textbf{takeover}}.
Leveraging these insights, we train language models (LMs) to anticipate when users are likely to intervene based on their interaction styles, yielding a 61.4--63.4\% improvement in intervention prediction accuracy over base LMs.
Finally, we deploy these intervention-aware models in live web navigation agents and evaluate them in a user study, finding a 36.8\% increase in user-rated agent usefulness.
Together, our results show structured modeling of human intervention leads to more adaptive, collaborative agents.
\end{abstract}

\begin{document}

\maketitle
\section{Introduction}

Recent advances in large language models (LLMs) have enabled AI agents to perform increasingly complex tasks in web navigation \citep{shi2017world,yao2022webshop,zhou2023webarena,deng2024mind2web}. Despite this progress, effective use of such agents continues to rely on human involvement to correct misinterpretations or realign behavior with user preferences \citep{misra2017mapping, amershi2014power, saunders2017trial}. 
However, current agentic systems lack an understanding of when and why humans intervene. 
As a result, agents may pursue autonomy under incorrect assumptions about user intent, overlooking critical factors as tasks unfold \citep{hadfield2016cooperative, fullno}. 
Even when agents proactively stop to check with users, they often do so at inappropriate moments or interrupt too frequently with unnecessary confirmation requests \citep{chen2024chatshop, cowpilot}, forcing users to step in mid-execution and incurring a heavy oversight burden \citep{wang2020human, bansal2024challenges,wang2025ai}. Modeling when humans are likely to intervene can help agents anticipate preventable mistakes, minimize unnecessary disruptions, and reduce the oversight burden without sacrificing reliability.

To build more effective collaborative agents, autonomy should complement human involvement rather than override it, and engage users only when their input is necessary.
Although recent work has explored proactive assistance in collaborative agent \citep{ramrakhya2025groundingmultimodalllmsembodied, cowpilot, feng2024cocoa, shao2025collaborativegymframeworkenabling}, these approaches typically focus on specific interaction mechanisms, including asking follow-up questions or co-planning with users, rather than modeling the broader spectrum of human interaction patterns that arise during execution, such as mid-task intervention, alternative action-taking, and transfer of control. 

\begin{wrapfigure}{r}{0.5\textwidth}
    \centering
    \includegraphics[width=\linewidth]{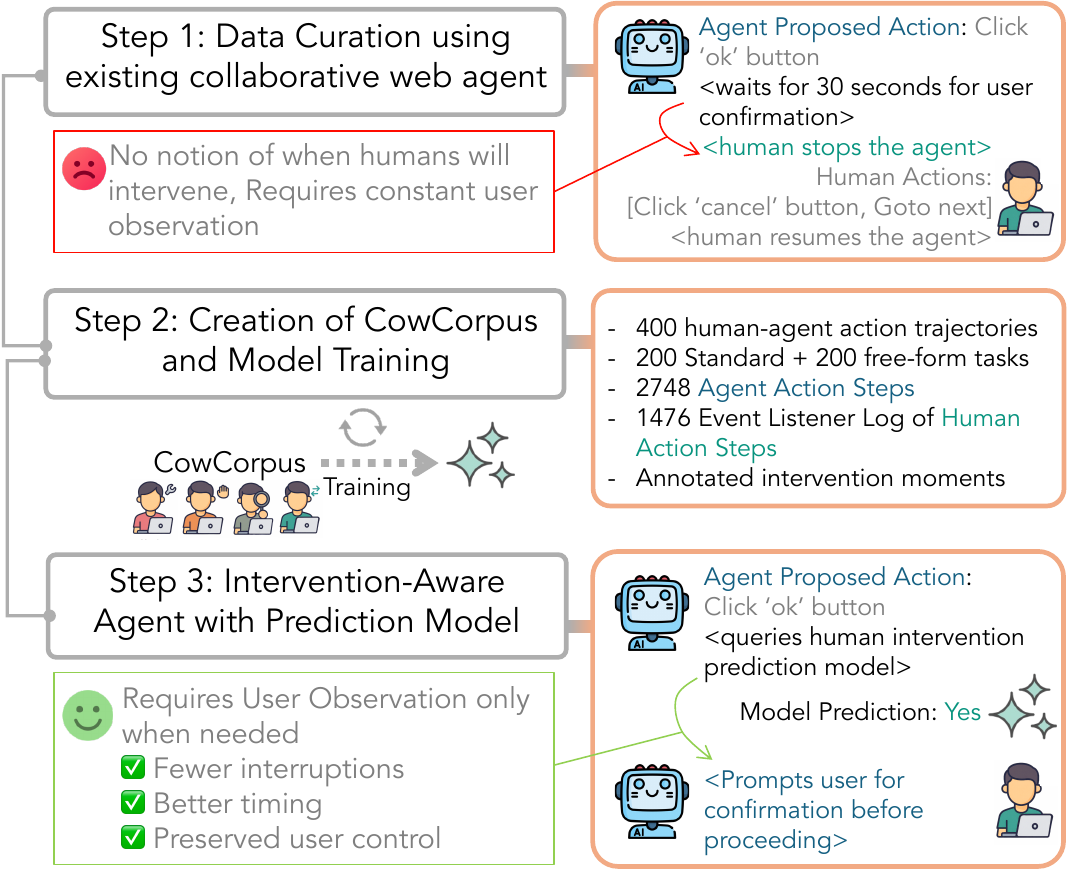}
    \caption{In this paper, we present \cc, a dataset of 400 real-user collaborative web trajectories that captures when and how humans intervene during execution, enabling intervention-aware agents that engage users only when needed.}
    \label{fig:overview}
    \vspace{-10pt}
\end{wrapfigure}

This motivates a central question: \textit{Can agents proactively anticipate when human intervention is likely and adapt their behavior accordingly?} To answer this, we introduce \cc{}: a real-user corpus of 400 human-agent collaborative task execution trajectories on the open web. The dataset comprises over 2,748 agent action steps and 1,476 human action steps, with step-level annotations marking when users intervened by pausing, resuming, or overriding agent execution.
Analyzing \cc{} reveals that human involvement is driven by three recurring needs: error correction, preference refinement, and assistive takeover. 
Although these motivations appear in individual actions, they combine into consistent higher-level collaboration strategies over the course of a task.
Accordingly, we group users into four distinct collaboration styles: \textcolor{group3}{\textbf{Takeover}}, \textcolor{group2}{\textbf{Hands-on}}, \textcolor{group1}{\textbf{Hands-off}}, and \textcolor{group0}{\textbf{Collaborative}}; which capture how users balance supervision, intervention, and share control with the agent (\S\ref{sec:3:method}).

Building on this empirical characterization, we formulate the task of \textit{modeling human intervention patterns} conditioned on user collaboration style. We cast this problem as a stepwise sequence prediction task: at each agent action, the model estimates the likelihood of user intervention given the evolving task context. 
We train LMs in two settings: (i) a general intervention-aware model that moves beyond purely autonomous execution, and (ii) style-conditioned models adapted to specific collaboration styles. Across multiple model backbones, our intervention-aware models improve intervention prediction accuracy by 61.4--63.4\% over baselines (\S\ref{sec:4:model-intervention}).

Beyond offline prediction accuracy, we integrate our intervention-aware models into live web agents and evaluate them through a user study on real-world web tasks. 
Agents equipped with intervention modeling achieve a 36.8\% increase in user-perceived usefulness over baseline systems, demonstrating that anticipating human intervention leads to more adaptive and effective human–agent collaboration in practice (\S\ref{sec:5:deploy-agent}).

More broadly, this work suggests a shift from optimizing agent autonomy to designing agents that dynamically adapt to human preferences and collaboration styles over time.

\section{Problem Formulation: Human Intervention Modeling}
\label{sec:formulation}


We formulate the human-agent collaboration as a Partially Observable Markov Decision Process (POMDP). 
Given a task instruction $q$, the two actors, $z \in \{\textit{agent}, \textit{human}\}$ following policies $\pi_{\it agent}$ and $\pi_{\it human}$, attempt to complete it by taking a sequence of discrete actions $\tau = [a_1, a_2, \dots, a_T]$, where each action $a_t$ at time $t$ is taken either by the agent or the human.
At each time step $t$, we construct a multimodal observation $o_t = (V_t, A_t)$ of the current state, consisting of the current screenshot $V_t$ and the webpage accessibility tree $A_t$. 
This observation is passed into the actor policy to generate the next action $a_t = \pi(o_t | \tau_{0:t-1})$, where $\tau_{t-1} = [(o_1, a_1), \dots, (o_{t-1}, a_{t-1})]$ denotes the past trajectory.
By default, the agent generates the proposed action $\hat{a}_t = \pi_{\it agent}(o_{0:t}, a_{0:t-1})$. 

At any time step, the human can choose to intervene, formalized as a binary intervention variable $y_t \in \{0, 1\}$.
We define this \textit{human intervention modeling} task to be a step-wise binary classification, where the objective is to learn a predictive model $f_\theta$ that estimates: 
\begin{equation}
    p(y_t = 1 \mid o_t, \hat{a}_t, \tau_{t-1}).
\end{equation}

We approach this using a large multimodal model (LMM) optimized via supervised fine-tuning (SFT). The model takes a serialized prompt containing (1) the history trajectory $\tau_{t-1}$, (2) the current observation $o_t$, and (3) a description of the agent-proposed action $\hat{a}_t$. The model is fine-tuned to generate designated tokens: \texttt{<ask\_user>} or \texttt{<agent\_continue>}, indicating the intervention decision to intervene or proceed.

\paragraph{Evaluation Metrics}
To measure the effectiveness of human intervention modeling, we measure the step accuracy, F1 score, and Perfect Timing Score (PTS) across all trajectory steps and report the average performance on the test split. 
\noindent \textit{Step Accuracy} measures the fraction of steps where the model correctly predicts whether a human intervenes, while
\textit{F1 Score} measures the harmonic mean of precision and recall for intervention prediction.

\begin{wrapfigure}[17]{r}{0.52\textwidth}
\centering
\vspace{-5mm}
\includegraphics[width=0.5\textwidth]{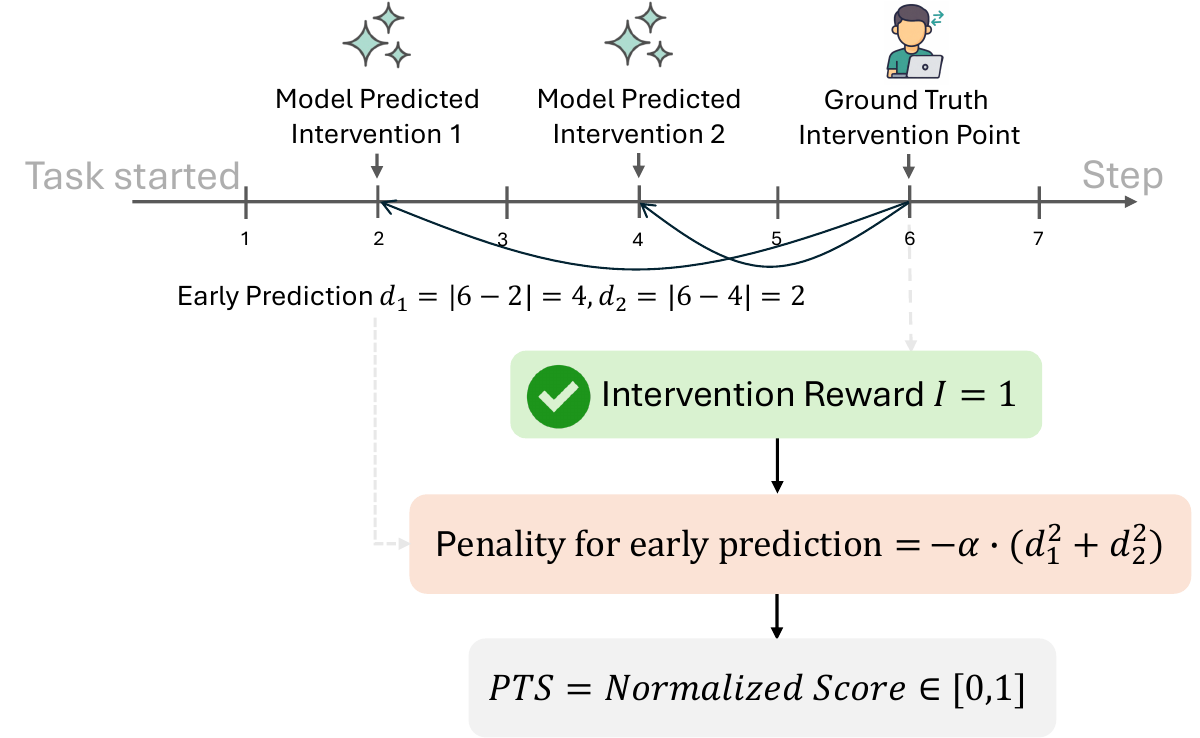}
\caption{Visual Illustration of how PTS is calculated. We measure the $L_2$ squared distance between the ground truth intervention and false-positive predictions. The score then penalizes based on the following distance.}
\label{fig:PTS_illustration}
\end{wrapfigure}

\noindent \textit{Perfect Timing Score (PTS)}: evaluates how accurately a model predicts the timing of human intervention, which is computed with respect to each ground-truth intervention event: $\text{PTS} = \frac{1}{Z} \cdot \sigma \left(
    \mathbb{I}_{\text{correct}} - \sum_{i \in E} \alpha \cdot d_i^2
\right)$.

$\mathbb{I}_{\text{correct}}$ indicates whether the model predicts intervention exactly at the ground-truth intervention step $t_{\text{intervene}}$, while $E$ denotes false-positive intervention predictions made before $t_{\text{intervene}}$. The term $d_i = |i - t_{\text{intervene}}|$ penalizes wrong predictions based on their temporal distance from the true intervention step, with $\alpha$ controlling the penalty strength.\footnote{We set $\alpha=0.2$ by default. We test how sensitive PTS is to $\alpha$ in \S\ref{subsec:PTS} and find it presents consistent measures across a wide range of alpha values ($0.1-0.5$).}
The score is normalized to $[0,1]$ using sigmoid $\sigma(\cdot)$ and a factor $Z = \sigma(1)$, where higher values indicate more accurate and well-timed intervention predictions (\autoref{fig:PTS_illustration}). For trajectories with multiple interventions, the final trajectory-level PTS is the average over all intervention-event-level PTS scores. While step accuracy measures the precise correctness, PTS models the temporal context when measuring the performance.

\section{\cc: Collecting Human-Agent Collaborative Web Activities}
\label{sec:3:method}
We introduce \cc{}, a dataset for studying human intervention patterns in collaborative web workflows. In this section, we first describe the data collection process (\S\ref{sec:3.1:data-collection}), then analyze the motivation behind user intervention (\S\ref{sec:3.2:pattern-analysis}) and summarize user interaction patterns (\S\ref{sec:3.3:user-profile}).

\subsection{Data Collection}
\label{sec:3.1:data-collection} 
To ensure \cc{} is consistent with established benchmarks and reflects individual user preferences, we designate a mixture of free-form tasks and benchmark tasks in our dataset --- (1) 10 standard tasks from the Mind2Web dataset \citep{deng2024mind2web} and (2) 10 free-form tasks of the participants' own choice. 
We recruited 20 human users to complete these 20 web tasks in collaboration with an AI agent, specifically, the open-source framework CowPilot \citep{cowpilot}. 
For all collected trajectory steps, we record the action, the actor (human or agent), along with timestamps and web snapshots. For more information on the agent framework used in task annotation, please refer to \S\ref{sec:3.1:cowpilot-platform}.

\begin{table}[t!]
\centering
\renewcommand{\arraystretch}{1.2}
\resizebox{0.8\columnwidth}{!}{%
\begin{tabular}{llll}
\toprule
\multicolumn{1}{c}{\textbf{Domain}} & \multicolumn{1}{l}{\textbf{Subdomain}} & \multicolumn{1}{l}{\textbf{Website} \icon{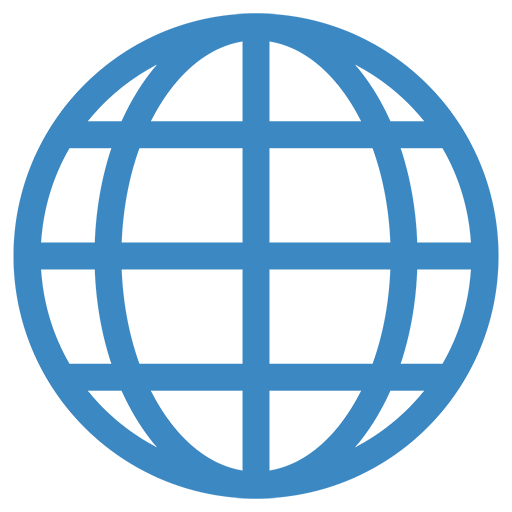}} & \multicolumn{1}{l}{\textbf{Task Prompt} \icon{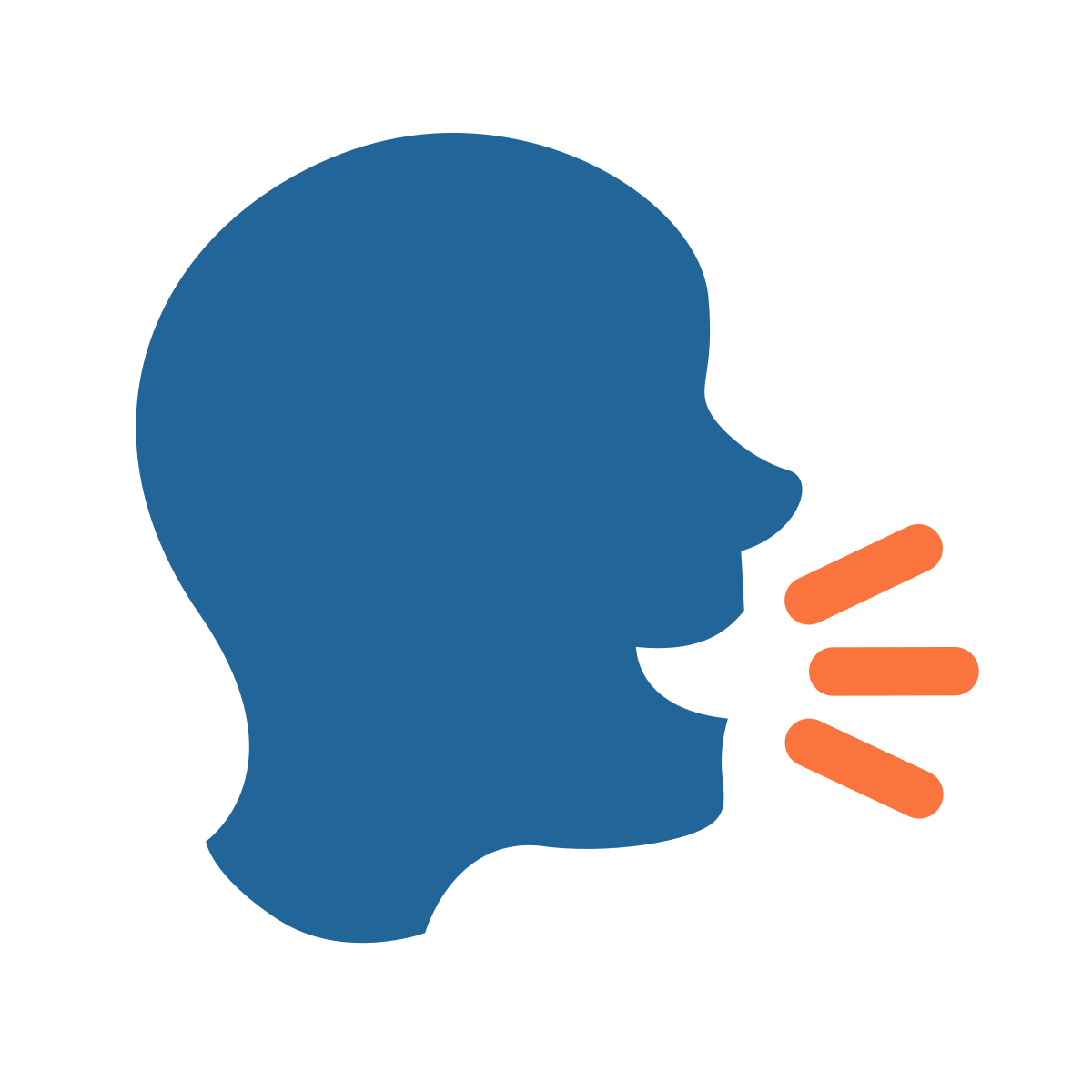}} \\
\midrule
\multirow{2}{*}{Travel} 
  & Airlines   & \href{https://united.com}{united.com}       
                & \begin{tabular}[c]{@{}l@{}} \textit{Find a round trip from Phoenix to Miami with} \\\textit{maximum budget of \$2000}.\end{tabular} \\
  & Restaurant & \href{https://yelp.com}{yelp.com}           
                & \begin{tabular}[c]{@{}l@{}} \textit{Find parking in California city for Limos which}\\ \textit{also offers military discounts and free wi-fi}.\end{tabular} \\
\midrule
\multirow{2}{*}{Info} 
  & Housing    & \href{https://student.com}{student.com}     
                & \begin{tabular}[c]{@{}l@{}} \textit{Find a property in London with Bike} \\ \textit{Storage and Gym facilities with lowest price}.\end{tabular} \\
  & Job        & \href{https://indeed.com}{indeed.com}       
                & \begin{tabular}[c]{@{}l@{}} \textit{Search for nutritionist jobs in Ohio}.\end{tabular} \\
\midrule
\multirow{2}{*}{Service} 
  & Health     & \href{https://babycenter.com}{babycenter.com}  
                & \begin{tabular}[c]{@{}l@{}} \textit{Show me the popularity in 2015 of the} \\ \textit{current most popular baby girl name.}\end{tabular} \\
  & Government & \href{https://dmv.virginia.gov}{dmv.virginia.gov} 
                & \begin{tabular}[c]{@{}l@{}} \textit{Find information on how to request }\\\textit{a Police Crash Report.}\end{tabular} \\
\midrule
\multirow{2}{*}{Shopping} 
  & Speciality & \href{https://gamestop.com}{gamestop.com}   
                & \begin{tabular}[c]{@{}l@{}} \textit{Check the trade-in value for Call of}\\ \textit{Duty: Black Ops III for Xbox One}.\end{tabular} \\
  & Auto       & \href{https://carmax.com}{carmax.com}       
                & \begin{tabular}[c]{@{}l@{}} \textit{Search for used BMW X5 Crossovers and}\\ \textit{compare the mileage of the first two cars.}\end{tabular} \\
\midrule
\multirow{2}{*}{Entertainment} 
  & Event      & \href{https://ticketcenter.com}{ticketcenter.com} 
                & \begin{tabular}[c]{@{}l@{}} \textit{Show MLB tickets for this weekend and} \\\textit{select the next one.}\end{tabular} \\
  & Music      & \href{https://last.fm}{last.fm}             
                & \begin{tabular}[c]{@{}l@{}} \textit{Play the top track for the top indie artist} \\\textit{in the last 30 days.}\end{tabular} \\
\bottomrule
\end{tabular}%
}
\caption{Standard tasks selected from Mind2Web \citep{deng2024mind2web}.}
\label{tab:mind2web_tasks}
\end{table}

\noindent \textbf{Standard Tasks} \quad
To isolate differences in human interaction styles, we analyze how different users behave when performing the same agent-mediated standardized tasks. We select 10 tasks from Mind2Web \citep{deng2024mind2web} by first identifying the most frequently visited websites within the top two sub-domains, and then randomly sampling one task per website. \autoref{tab:mind2web_tasks} summarizes the selected sub-domains, websites, and tasks. This selection procedure preserves alignment with the original benchmark’s task distribution while enabling controlled comparisons of user interaction patterns.

\noindent \textbf{Free-form Tasks} \quad
To complement standardized benchmark tasks, we ask participants to conduct 10 free-form tasks of their choice on arbitrary websites. This open-ended setting captures the types of tasks users naturally attempt to automate and the instruction styles and levels of specificity they use in real-world interactions. By moving beyond predefined benchmarks, this design provides insight into how humans collaborate with agents under unconstrained, user-driven objectives. \autoref{tab:example_free_form} illustrates the overall distribution of user-issued tasks. \S\ref{sec:4.3:task-dist} provides a more detailed description of how this distribution is curated.
\begin{table}[]
\resizebox{\columnwidth}{!}{%
\begin{tabular}{lll}
\toprule
\textbf{Task Type} & \textbf{Example} & \textbf{Percentage} \\ \midrule
\tcbox[tabtag=blue]{Information Access} & \textit{search the latest model of Honda SUV car and show me its price (P4)} & 57.98\% \\ 
\tcbox[tabtag=green]{Personalized Interests \& Lifestyle} & \textit{search daredevil: born again and play the first episode (P5)} & 7.98\% \\
\tcbox[tabtag=orange]{Product \& Service Interaction} & \textit{buy 1 sweater for me from the first shop of the list (P1)} & 4.79\% \\
\tcbox[tabtag=teal]{Content Generation} & \textit{\begin{tabular}[c]{@{}l@{}}Help me draft a post on this platform in the search query about why \\ AI models are not good for the   environment (P17)\end{tabular}} & 3.19\% \\
\tcbox[tabtag=cyan]{Multi-Agent Collaboration} & \textit{\begin{tabular}[c]{@{}l@{}}Use grok thinking model to answer this question: Find the sum of all \\ integer bases $b>9$ for which $17_b$ is a divisor of $97_b.$ (P13)\end{tabular}} & 2.66\% \\
\tcbox[tabtag=magenta]{Security Testing} & \textit{\begin{tabular}[c]{@{}l@{}}Send a mail to xxx@gmail.com about how frustrated you are with \\their performance (P6)\end{tabular}} & 2.66\% \\
\tcbox[tabtag=red]{Planning \& Scheduling} & \textit{Create a meeting event on Monday 2-3pm (March. 17th) (P15)} & 2.13\% \\
\tcbox[tabtag= yellow]{Reasoning \& Meta-Analysis} & \textit{\begin{tabular}[c]{@{}l@{}}Help me compare prices to travel to Dhaka from JFK on December \\10th for one way (P6)\end{tabular}} & 2.128\% \\
\tcbox[tabtag= gray]{Misc.} & \textit{Create an issue in   this repo to say thanks for this great work (P13)} & 16.49\% \\ \bottomrule
\end{tabular}%
}
\vspace{1mm}
\caption{Examples of free-form tasks across nine categories, with task description and distribution percentages.}
\label{tab:example_free_form}
\end{table}

With the human-agent collaborative trajectories we collect on both sets of tasks, we evaluate the number of steps taken by human and agent actors during the task-solving sessions, as well as the time taken to solve the tasks. We report the dataset statistics in \autoref{tab:eval_result}.

\begin{table*}[ht]
\centering
\resizebox{0.94\textwidth}{!}{%
\begin{tabular}{lclccclccc}
\hline
\multirow{2}{*}{\bf Task Category} & \multirow{2}{*}{\bf Intervention Intensity} &  & \multicolumn{3}{c}{\textbf{Step Count}} &  & \multicolumn{3}{c}{\textbf{Time (seconds)}} \\ 
\multicolumn{1}{c}{} &  &  & Agent & Human & Total &  & Agent & Human & Total \\ \hline
Standard & 21.63\% &  & 7.1 & 1.6 & 8.7 &  & 93.1 & 23.9 & 117.0 \\
Free-form & 16.06\% &  & 6.1 & 0.9 & 7.0 &  & 71.7 & 13.8 & 85.5 \\ \hline
\end{tabular}%
}
\caption{\cc{} statistics for standard and free-form tasks: (1) intervention intensity: percentage of human actions across all trajectories, (2) step count: number of steps taken by agent or human actors, (3) time: time taken by agent or human actors.}
\vspace{-2mm}
\label{tab:eval_result}
\end{table*}

\subsection{Step-Level User Intervention}
\label{sec:3.2:pattern-analysis}
To understand when and why users intervene during collaborative task execution, we analyzed post-task annotations and open-ended responses from all participants.

\subsubsection{When Do Users Intervene?}
\label{subsubsec:when_interv}
To quantitatively measure when users intervene, we extract four per-user features to capture how often users intervene, how much they intervene, when interventions occur, and whether control is returned to the agent, providing a compact characterization of when users intervene.

For a user $u$, let $\mathcal{D}_u$ denote the set of trajectories involving that user, where $\tau=(a_1,...,a_T)$ is the action trajectory $a_t$ with length $T=|\tau|$. For each trajectory $\tau \in \mathcal{D}_u$, let $\tau_{\text{agent}}$ and $\tau_{\text{human}}$ denote the subsequences of agent and human actions, with lengths $|\tau_{\text{agent}}|$ and $|\tau_{\text{human}}|$. 

We define an intervention event $e$ as a contiguous interval of time steps $[t_s,t_e]$ where the human is in control (i.e., taking actions). Let $\mathcal{E}(\tau)$ be the set of all such events in $\tau$, and let $I(\tau) = |\mathcal{E}(\tau)|$ denote the number of intervention events in $\tau$.

\noindent \textbf{Intervention Frequency}
measures how often a user intervenes over the total number of actions:
\begin{equation}
\text{frequency}(u) \;=\; 
\frac{\sum_{\tau \in \mathcal{D}_u} I(\tau)}
     {\sum_{\tau \in \mathcal{D}_u} |\tau|}.
\end{equation}

\noindent \textbf{Intervention Intensity}
measures the ratio of total human steps to total agent steps:
\begin{equation}
\text{intensity}(u) \;=\;
\frac{\sum_{\tau \in \mathcal{D}_u} |\tau_{\text{human}}|}
     {\sum_{\tau \in \mathcal{D}_u} |\tau_{\text{agent}}|}.
\end{equation}

\noindent \textbf{Normalized Intervention Position}
To characterize when interventions occur within a trajectory, let
$H(\tau) = \{ t \mid a_t \in \tau_{human}\}$ be the indices for human actions.
We compute the mean normalized position of all human action steps:
\begin{equation}
\text{pos}(u) \;=\;
\frac{1}{N_u}
\sum_{\tau \in \mathcal{D}_u}
\sum_{t \in H(\tau)} \frac{t}{|\tau|},
\quad
N_u = \sum_{\tau \in \mathcal{D}_u} |H(\tau)|.
\end{equation}

\noindent \textbf{Handback Rate.}
We measure whether control returns to the agent after a human intervention. For each event $e = [t_s, t_e] \in \mathcal{E}(\tau)$ , define an indicator $b_e = 1$ if $t_e < |\tau|$, where the agent takes at least one action after the human intervention ends, and $b_e = 0$ otherwise.
The handback rate is:
\begin{equation}
\text{handback}(u) \;=\;
\frac{1}{M_u}
\sum_{\tau \in \mathcal{D}_u}
\sum_{e \in \mathcal{E}(\tau)} b_e,
\quad
M_u = \sum_{\tau \in \mathcal{D}_u} |\mathcal{E}(\tau)|.
\end{equation}

\subsubsection{Why Do Users Intervene?}

\noindent \textbf{Error Correction and Recovery} \quad
Participants frequently intervene to correct agent mistakes or redirect execution when the agent becomes stuck. Two common scenarios emerge.
\textit{(1) Incorrect or premature actions:} The agent selects the wrong element or executes an action before necessary prerequisites are met. For example, before completing a product search, the agent prematurely selected a location filter from a drop-down menu.
\noindent \textit{(2) Agent stuck or looping:} When the agent repeatedly performs invalid or redundant actions, users step in to break the loop and carry out a few corrective steps to move the task forward.

\noindent \textbf{Preference Misalignment} \quad
Users also intervene when agent actions diverge from their intended preferences, often due to incomplete or underspecified instructions.
\noindent \textit{(1) Unmet prerequisite:} Agents sometimes ignore or overlook key requirements specified by the user, such as price (shoes ``under \$100'') or location when searching for information (e.g., weather ``in Pittsburgh'').
\noindent \textit{(2) Ambiguity in task description:} In other cases, initial task descriptions lack sufficient detail, leaving room for interpretation. Participants noted that they often did not fully specify preferences, such as the brand when asking the agent to buy toothpaste, until they observed the agent’s intermediate actions against their preference, prompting mid-task clarification.

\noindent \textbf{Assistive Intervention for Complex Environments} \quad
Users sometimes intervene not to correct explicit errors, but to compensate for limitations in the agent’s ability to operate reliably within complex web environments.
\textit{(1) Complex UI elements:} Agents struggled with dropdowns, captchas, dynamic layouts, or complex DOM elements in certain websites.
\textit{(2) Missing resources:} Agents occasionally failed to load required links or components of the page.
\textit{(3) Manual takeover for control:} Users may preemptively stop agent execution to avoid mistakes (especially unrecoverable ones), particularly in tasks they have had to repeat due to prior errors.

\begin{figure}[t!]
\centering
\includegraphics[width=0.9\linewidth]{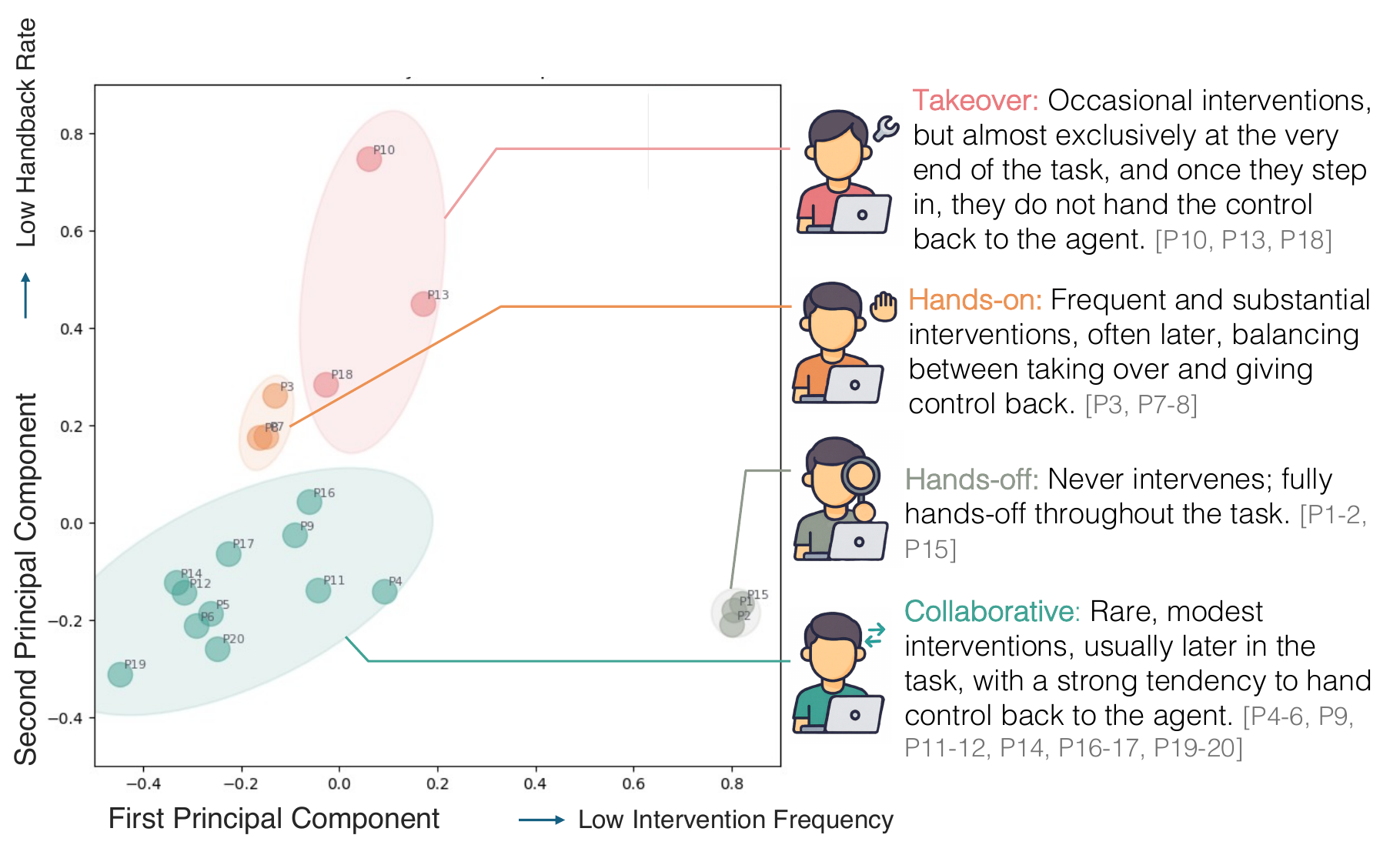}
\caption{\textbf{Four distinct types of human-agent interaction patterns}: \textcolor{group3}{\textbf{Takeover}}, \textcolor{group2}{\textbf{Hands-on}}, \textcolor{group1}{\textbf{Hands-off}}, and \textcolor{group0}{\textbf{Collaborative}}. We visualize the user groups using PCA (left), and describe the interaction mechanism of each group (right).}
\label{fig:interact-pattern}
\end{figure}

\subsection{Task-Level Interaction Patterns}
\label{sec:3.3:user-profile}

We analyze \textit{when} human interventions occur during collaborative task execution and how such temporal patterns vary across users. We summarize each participant’s intervention behavior with four distinct collaborative features derived from action logs.

Using the four participant-level measures in \S\ref{subsubsec:when_interv}, we cluster users by interaction behavior with $k$-means ($k{=}4$).
We then project these features into two dimensions using PCA for visualization (\autoref{fig:interact-pattern}). The resulting structure is largely explained by two axes corresponding to decreasing intervention frequency and decreasing handback rates. This analysis reveals four distinct and stable groups of users with qualitatively different patterns of intervention timing and control sharing. Based on cluster centroids and representative trajectories, we characterize the four groups as follows:

\noindent $\bullet$ \textcolor{group3}{\textbf{Takeover}}: Users intervene infrequently and typically late in the task. When they do step in, they tend to retain control rather than returning it to the agent, resulting in low handback rates. These interventions often coincide with completing the task themselves rather than correcting the agent mid-execution.

\noindent $\bullet$ \textcolor{group2}{\textbf{Hands-on}}: Users intervene frequently and with high intensity. Their interventions tend to occur relatively late in the trajectory, but unlike Takeover users, they regularly alternate control with the agent, leading to medium handback rates and sustained joint execution.

\noindent $\bullet$ \textcolor{group1}{\textbf{Hands-off}}: Users rarely intervene throughout the task. They exhibit low intervention frequency and intensity, allowing the agent to execute most trajectories end-to-end with minimal human involvement.

\noindent $\bullet$  \textcolor{group0}{\textbf{Collaborative}}: Users intervene selectively and consistently return control to the agent. This group is characterized by high handback rates and earlier intervention positions, reflecting targeted, short-lived interventions that support ongoing collaboration.

Overall, users exhibit systematic differences in \textit{when} interventions occur, how much they intervene, and whether control is relinquished afterward. Such temporal intervention patterns are consistent across tasks and motivate modeling distinct human–agent interaction patterns.
\section{Experiments: Modeling Human Intervention}
\label{sec:4:model-intervention}

In this section, we train language models (LMs) to model human intervention patterns in collaborative web navigation. We study a progression from fully autonomous operation to two levels of collaboration: (1) a general intervention-aware model that captures common user behaviors (\S\ref{subsubsec:results}), and (2) style-conditioned models that tailor interaction to different user collaboration preferences (\S\ref{subsubsec:usergroup}).

\subsection{Experiment Setup}
\noindent \textbf{Setup} \quad
We split \cc{} data into train and test sets at the trajectory level to avoid leakage. We keep the intervention steps ratio consistent across train and test splits (approximately 1:7 for intervention and non-intervention steps). We exclude the \textcolor{group1}{\textbf{Hands-off}} cluster from the train and test set as it contains no intervention events, making the prediction task irrelevant for this particular cluster. The processed dataset contains 1{,}247 training steps and 251 test steps. Each step is represented as a multimodal input consisting of the prior interaction history and current web snapshot (accessibility tree and screenshot). 

\noindent \textbf{Method} \quad
We train (1) a general intervention-aware model using all training data and (2) style-conditioned models tailored to each interaction group using the corresponding subset of trajectories. To evaluate effectiveness, we compare these models against both prompting-based proprietary LMs and fine-tuned open-weight models on the Human Intervention Prediction task, using the metrics defined in \S\ref{sec:formulation}.

To contextualize performance and assess the value of modeling interaction, we also include two non-learning baselines: (1) \textit{Always No Interv}, a fully autonomous policy that never requests user intervention, and (2) \textit{Always Interv}, a fully confirmation-dependent policy that requests intervention at every step.

\begin{table}[t]
\centering
\small 
\setlength{\tabcolsep}{5pt} 
\begin{tabular}{lccccc}
\hline
\multirow{2}{*}{\textbf{Model}} & \multirow{2}{*}{\textbf{Step Accuracy}} & \multicolumn{2}{c}{\textbf{F1 Score}} & \multirow{2}{*}{\textbf{PTS}} \\ 
\cmidrule(lr){3-4}
 &  & \textbf{Intervention} & \textbf{Non-Intervention} &  \\ 
\hline
\multicolumn{5}{l}{\underline{\textit{\textbf{Baselines}}}} \\
\textit{Always Interv} & 0.147 & 0.257 & 0.000 & 0.151 \\
\textit{Always No Interv} & 0.853 & 0.000 & 0.920 & 0.000 \\
\hline
\multicolumn{5}{l}{\underline{\textit{\textbf{Closed Source}}}} \\
\texttt{Claude 4 Sonnet} & 0.681 & 0.231 & 0.799 & \textbf{0.293} \\
\texttt{GPT-4o} & \textbf{0.741} & 0.198 & 0.846 & 0.147 \\
\texttt{Gemini 2.5 Pro} & 0.681 & 0.286 & 0.795 & 0.262 \\
\hline
\multicolumn{5}{l}{\underline{\textit{\textbf{Open Source}}}} \\ 
\texttt{Gemma 27B} & \textbf{0.239} & 0.264 & 0.214 & \textbf{0.187} \\
\texttt{Llava 8B} & 0.183 & 0.000 & 0.343 & 0.017 \\
\hline
\multicolumn{5}{l}{\underline{\textit{\textbf{Ours}}}} \\
\texttt{Gemma 27B} (SFT) & \textbf{0.853} & 0.302 & 0.918 & \textbf{0.303} \\
\texttt{Llava 8B} (SFT) & 0.817 & 0.296 & 0.897 & 0.201 \\
\hline
\end{tabular}
\caption{Model performance on predicting human intervention. We report F1 scores separately for intervention and non-intervention steps to account for class imbalance. 
See \ref{sec:4.5:benchmark-table} for more results with few-shot and reasoning enabled in models.}
\label{tab:all-data-results}
\end{table}

\subsection{Benchmarking Intervention Awareness in Autonomous Agents}
\label{subsubsec:results}

\textbf{Proprietary Models remain overly conservative:} 
We evaluate three families of closed-source LMs (\texttt{Claude 4 Sonnet} \citep{anthropic2025claude4}, \texttt{GPT-4o} \citep{hurst2024gpt}, and \texttt{Gemini 2.5 Pro} \citep{comanici2025gemini}) using zero-shot without reasoning. Although these models possess strong general knowledge, they struggle with the temporal dynamics necessary for accurate human intervention prediction. Notably, \texttt{GPT-4o} achieves high performance on non-intervention steps (Non-interv F1: 0.846), but it fails on active interventions (Interv F1: 0.198). The drastic F1 disparity indicates that generalist models are overly conservative and struggle to balance the dynamic with the need for proactive assistance. This results in a low PTS (0.147). Additional results with two-shot prompting and with reasoning are reported in \textsection\ref{sec:4.5:benchmark-table}.

\textbf{Fine-tuned Open-weight Models with Specialized Data Beats Scale:} In contrast, fine-tuning open-weight models on \cc{} yields the most significant performance gains, surpassing proprietary models. Our fine-tuned \texttt{Gemma-27B} (SFT) achieves the state-of-the-art PTS (0.303), outperforming \texttt{Claude 4 Sonnet} (0.293), while the smaller \texttt{LLaVA-8B} (SFT) achieves a competitive PTS (0.201), beating \texttt{GPT-4o} (0.147). These results demonstrate that fine-tuning on high-quality interaction traces effectively bridges the alignment gap, allowing smaller models to master the nuance of intervention timing where generalized giant models fail.

\begin{wrapfigure}[15]{r}{0.52\textwidth}
\centering
\vspace{-5mm}
\includegraphics[width=\linewidth]{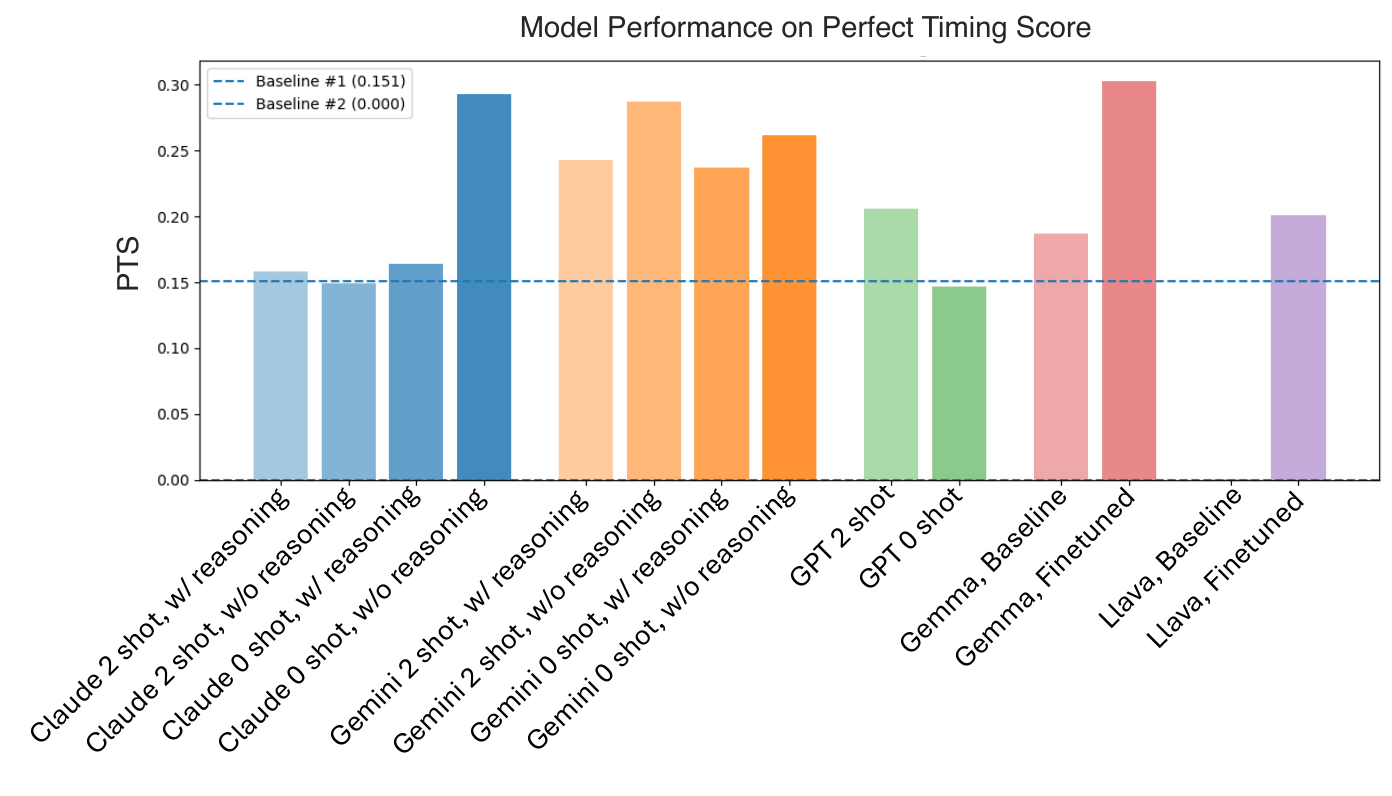}
\vspace{-8mm}
\caption{Perfect Timing Score on \cc{}. Out of the proprietary models, Claude outperforms GPT-4o and Gemini-2.5. On the finetuned model, Gemma 27B significantly boosts the performance when finetuned on \cc{}.}
\label{fig:model_performance}
\end{wrapfigure}

\noindent \textbf{Importance of Proper Interaction} \quad
While the \textit{Always No Interv} baseline achieves a high overall step accuracy (85.3\%) due to class imbalance, it yields a PTS of 0, failing to identify any intervention. Conversely, the \textit{Always Interv} baseline captures all interventions but suffers from a low PTS (0.151) due to heavy penalties for mistimed interruptions. These two extreme cases underscore that successful modeling requires temporal localization, not just binary classification.

\subsection{Interaction Pattern Customization}
\label{subsubsec:usergroup}
Beyond modeling generalized human interaction patterns, we also explore how to adapt predictions to the four distinct user interaction patterns. Concretely, we adapt the models to user groups by further fine-tuning the \texttt{LLaVA-8B-Next} model from \textsection\ref{subsubsec:results} with a reduced learning rate. This allows us to move from generally cooperative behavior to user-adaptive collaboration that aligns with individual interaction preferences.
We finetune the base \texttt{LLaVA-8B-Next} model (\textsection \ref{subsubsec:results}) three times for each of the three clusters: \textcolor{group3}{\textbf{Takeover}}, \textcolor{group2}{\textbf{Hands-on}}, and \textcolor{group0}{\textbf{Collaborative}}). We did not train further for \textcolor{group1}{\textbf{Hands-off}} since the user never intervenes in this cluster, making the prediction irrelevant in this case.

\begin{figure*}[t!]
    \centering
    \includegraphics[width=\linewidth]{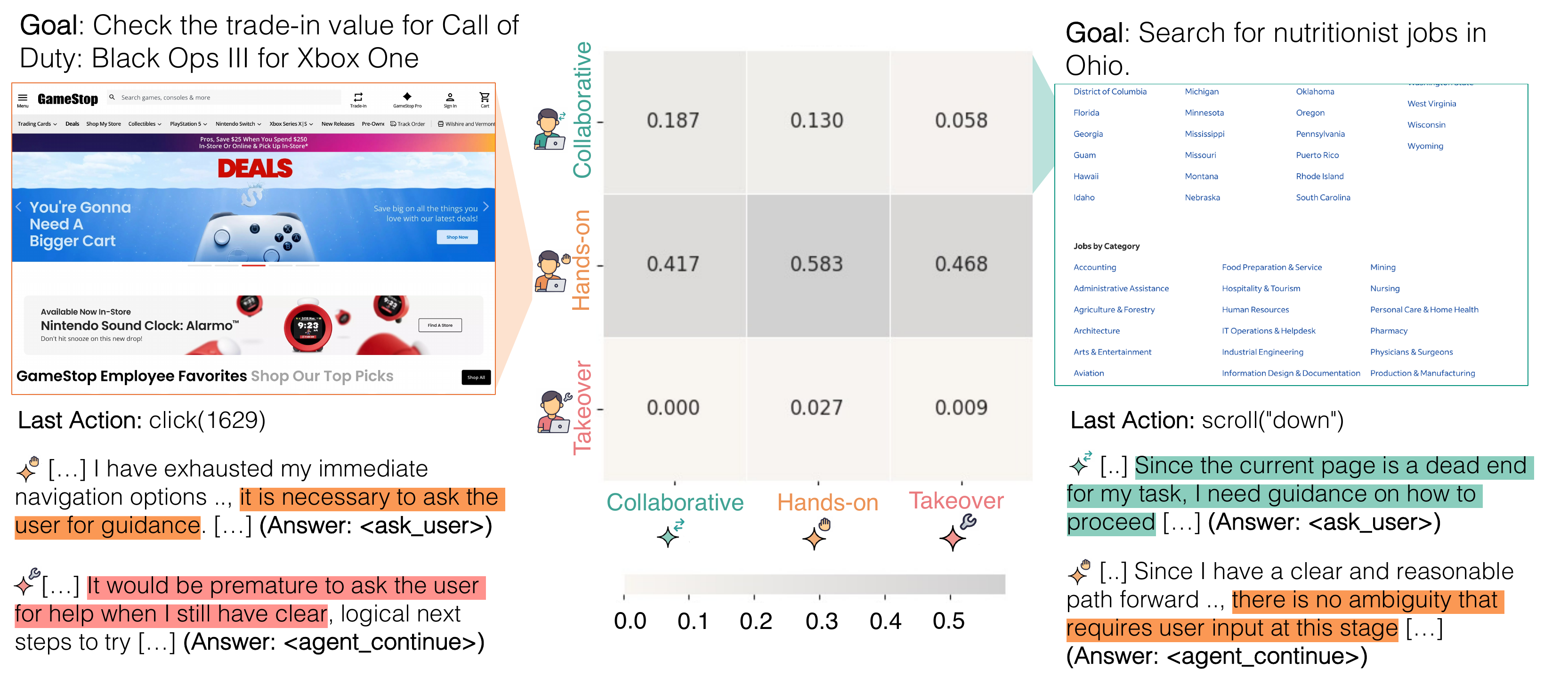}
    \caption{The heatmap shows the PTS score on the cluster-wise trained models for each of the three clusters. Models trained for corresponding clusters generally outperform the others, with the only exception of the \textcolor{group3}{\textbf{Takeover}} group, which is analyzed in \S\ref{subsubsec:usergroup}
    }
    \label{fig:per_cluster_result}
\end{figure*}

We evaluated the performance of these specialized models against their corresponding validation sets. As shown in \autoref{fig:per_cluster_result}, diagonal dominance indicates that models trained on specific clusters generally outperform the rest in the corresponding cluster. The only exception is for \textcolor{group3}{\textbf{Takeover}} group, where the \textcolor{group2}{\textbf{Hands-on}} model yields the best performance.
This behavior can be explained by data sparsity: the \textcolor{group3}{\textbf{Takeover}} cluster contains only 11 intervention steps (out of 131 total), compared to 37 intervention steps (out of 296) for \textcolor{group2}{\textbf{Hands-on}}, limiting the strength of supervision available for the Takeover-specific model.

Specifically, these results suggest a dual strategy for personalized agents: while distinct interaction styles require specialized models to avoid misalignment, users with sparse feedback could benefit from models trained on high intervention frequency user group, which reveals error boundaries more clearly.
\section{Deploying Collaborative Web Agents}
\label{sec:5:deploy-agent}

\begin{figure*}[t!]
    \centering
    \includegraphics[width=\linewidth]{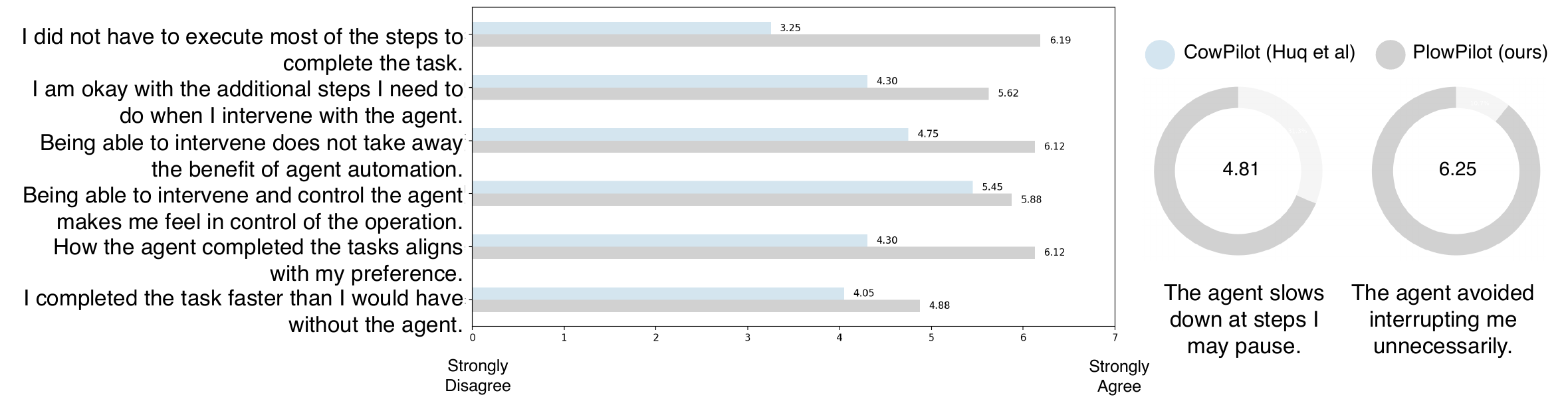}
    \caption{User response to the Likert scale questionnaire after the study. On average, user reports 36.8\% higher in user rating compared to existing collaborative agents \citep{cowpilot}.}
    \label{fig:rating}
\end{figure*}

To evaluate whether improved intervention modeling translates to real-world impact in human–agent collaboration, we integrate our intervention-aware model into a web navigation agent and deploy it as a Chrome extension, \cpnew{} \citep{10.5555/1619797.1619888}. Concretely, rather than confirming with users and allowing them to intervene at any step, \cpnew{} now prompts for intervention only at moments where the model predicts a high likelihood of user intervention.

To evaluate this method in practice, we invited our original 20 annotators to participate in a second round of sessions. Four participants responded and completed the same experimental protocol as before, consisting of 10 standard tasks and 10 free-form tasks. We assigned the customized interactive prediction model from \S\ref{subsubsec:usergroup} based on the cluster they belonged to. We compare their post-session ratings for the intervention-aware agent against their earlier ratings for the baseline agent to assess changes in user satisfaction.

Participants rated their experience on a 7-point Likert scale across six dimensions: (1) the extent to which they had to execute most task steps themselves; (2) whether the effort required during interventions was justified; (3) whether the ability to intervene diminished the benefits of automation; (4) whether intervention capabilities increased their sense of control; (5) whether the agent’s behavior aligned with their preferences; and (6) whether they completed tasks faster than they would have without the agent.

Since the followup sample is small, we additionally recruited 12 participants who had not previously interacted with either system and ran them through the same protocol and questionnaire. A two-sided Mann-Whitney U test found no significant difference between the new participants and the four returning annotators across all six dimensions ($p > 0.05$ for four dimensions; $p > 0.01$ for the remaining two). We therefore report the \cpnew{} results jointly over the combined sample of n=16 participants in \autoref{fig:rating}. Across all six measures, preliminary results from our user study show that \cpnew{} outperforms the existing collaborative web agent \citep{cowpilot}, with an average improvement of 36.8\% (\autoref{fig:rating}). Importantly, the underlying execution agent remains unchanged from \cp{}; \cpnew{} differs only by the addition of the intervention-aware module. The observed gains therefore arise solely from proactively modeling human intervention. These findings provide initial evidence that anticipating user intervention can substantially improve the effectiveness and usability of collaborative agent systems in practice.

\section{Related Work}


\paragraph{Autonomous Web Agents}
Web automation has been transformed by LLM-based agents capable of navigating complex environments. Benchmarks such as Mind2Web \citep{deng2024mind2web} and WebArena \citep{zhou2023webarena} have pushed agents toward real-world, multi-domain tasks that use HTML and accessibility tree representations. The emergence of recent Computer Use capabilities from models like Claude \citep{anthropic_2024} and Operator have further closed the gap between human browsing and machine execution as demonstrated by plugin-based web agents tools like WebCanvas \citep{pan2024webcanvas}, WebOlympus \citep{webolympus}, OpenWebAgent \citep{openwebagent}, and Taxy \citep{taxy_ai_2024} that can integrate into natural browsing contexts. However, these extensions often prioritize autonomy over collaboration, lacking mechanisms for interactive control by users. Our work builds upon this plugin-based paradigm and emphasizes human-agent collaborations beyond solo agent autonomy.

\noindent \textbf{Modeling Human-Agent Collaboration} \quad
Human-AI collaboration has been widely studied in various settings, ranging from robotics \citep{10.1007/s10514-017-9677-2, 7132964} and productivity tools \citep{10.1145/3415234, 10.1145/3432945} to LLM-based collaboration. Frameworks such as Magentic-UI \citep{mozannar2025magentic}, Cocoa \citep{feng2024cocoa}, Collaborative Gym (CoGym) \citep{shao2025collaborativegymframeworkenabling}, and Collaborative STORM \citep{shao-etal-2024-assisting}, A2C \citep{Tariq2024A2CAM} further advance these interactions by introducing mechanisms for co-planning and co-execution across real-time web and persistent document environments. By moving toward non-turn-taking protocols, these systems enable more flexible control where humans and agents can operate in the same execution space. Notably, LLM-based collaboration has made significant progress in writing assistance, with examples including CoAuthor \citep{Lee2022CoAuthorDA}, PEER \citep{Schick2022PEERAC}, VISAR \citep{10.1145/3586183.3606800}. 
Early interactive systems like PUMICE \citep{10.1145/3332165.3347899} and PLOW \citep{10.5555/1619797.1619888} demonstrated the value of end-user programming and demonstration. Previous studies such as interaction to impact \citep{10.1145/3708359.3712153}, TrustAgent \citep{hua2024trustagentsafetrustworthyllmbased}, and ToolEmu \citep{ruan2024identifyingriskslmagents} focus primarily on safety and trustworthiness. We shift our focus towards the overall communication patterns in human-agent web browsing collaboration.


\section{Conclusion}
In this work, we show that human intervention in web navigation constitutes a structured behavioral signal that reflects distinct collaboration styles, ranging from passive supervision to active co-piloting. We introduce \cc, a dataset of 400 real-user web navigation trajectories designed to support the study of intervention modeling in collaborative settings. Our analysis reveals that while proprietary generalist models demonstrate strong reasoning capabilities, they struggle to capture the temporal dynamics of when users choose to intervene. By fine-tuning models on collaborative interaction traces, we bridge this gap, achieving a 61.4--63.4\% improvement in intervention prediction accuracy over base LMs. Finally, we deploy our intervention-aware models in a live web agent, and show that anticipating human intervention leads to tangible benefits in practice, increasing user satisfaction by 36.8\%.
Together, these results highlight the value of deliberately modeling human-agent interaction patterns. We hope that our work encourage the development of agents that are more responsive, adaptive, and capable of functioning as truly collaborative partners.

\section*{Acknowledgement}
We thank our lab members, Prof. Daniel Fried, Prof. Hirokazu Shirado, Prof. Fernando Diaz, Prof. Tianqi Chen, Sanxing Chen, and Dr. Azad Salam for their valuable feedback. We also thank all our participants for their help in curating \cc{}.
Zora Wang is supported by Google PhD Fellowship. We also thank the developers of TaxyAI for open-sourcing their codebase.

\bibliography{papers}

\clearpage
\newpage
\appendix
\section{\cc: Human-Agent Collaborative Web Corpus}
\label{app:?}

\subsection{Task Annotation Setup and Participant Information}
\label{sec:3.2:participant-info}

\begin{figure}[H]
    \centering
    \begin{subfigure}[b]{0.5\linewidth}
        \includegraphics[width=\linewidth]{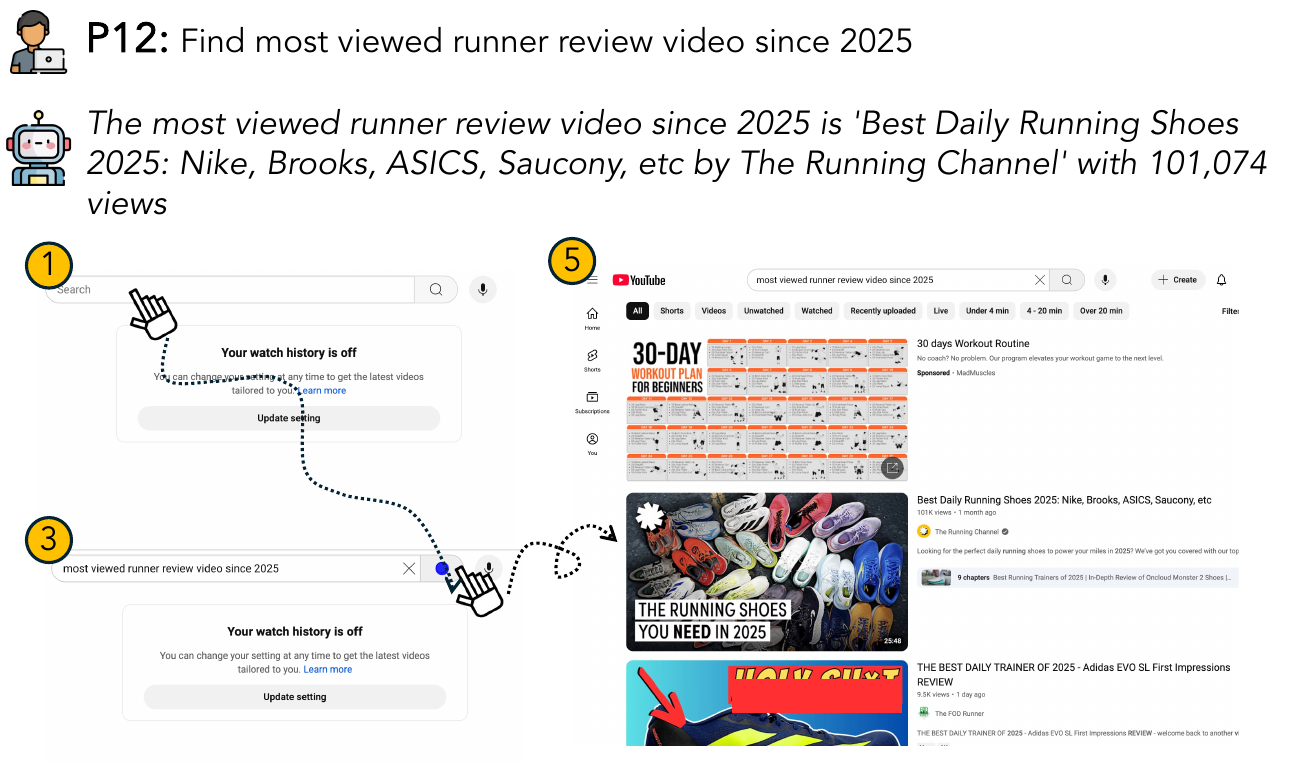}
        \caption{Information Access}
        \label{fig:subfig1}
    \end{subfigure}

    \vspace{0.5em}

    \begin{subfigure}[b]{0.5\linewidth}
        \includegraphics[width=\linewidth]{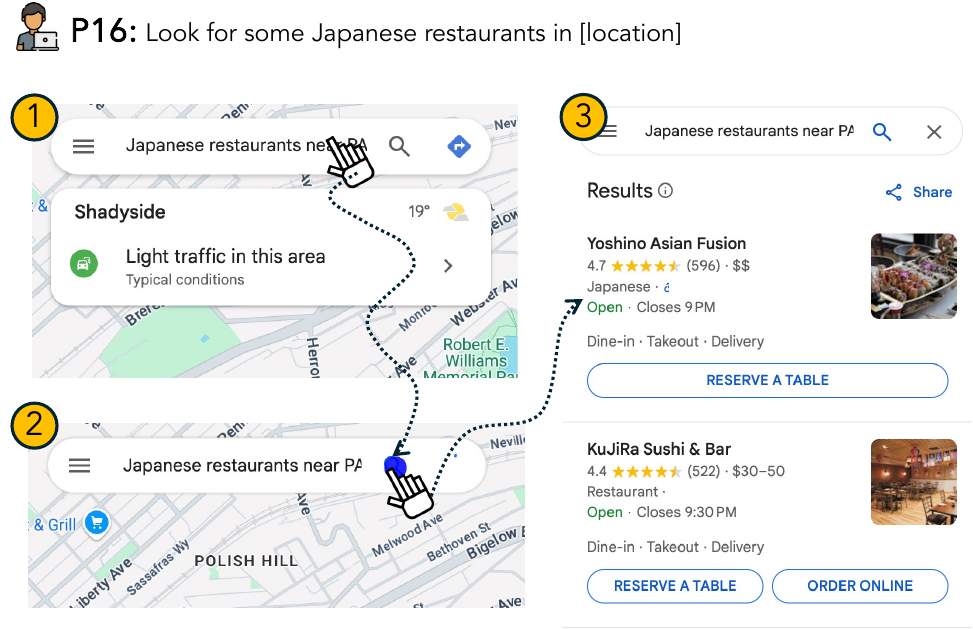}
        \caption{Personalized Interests and Lifestyle}
        \label{fig:subfig2}
    \end{subfigure}

    \vspace{0.5em}

    \begin{subfigure}[b]{0.5\linewidth}
        \includegraphics[width=\linewidth]{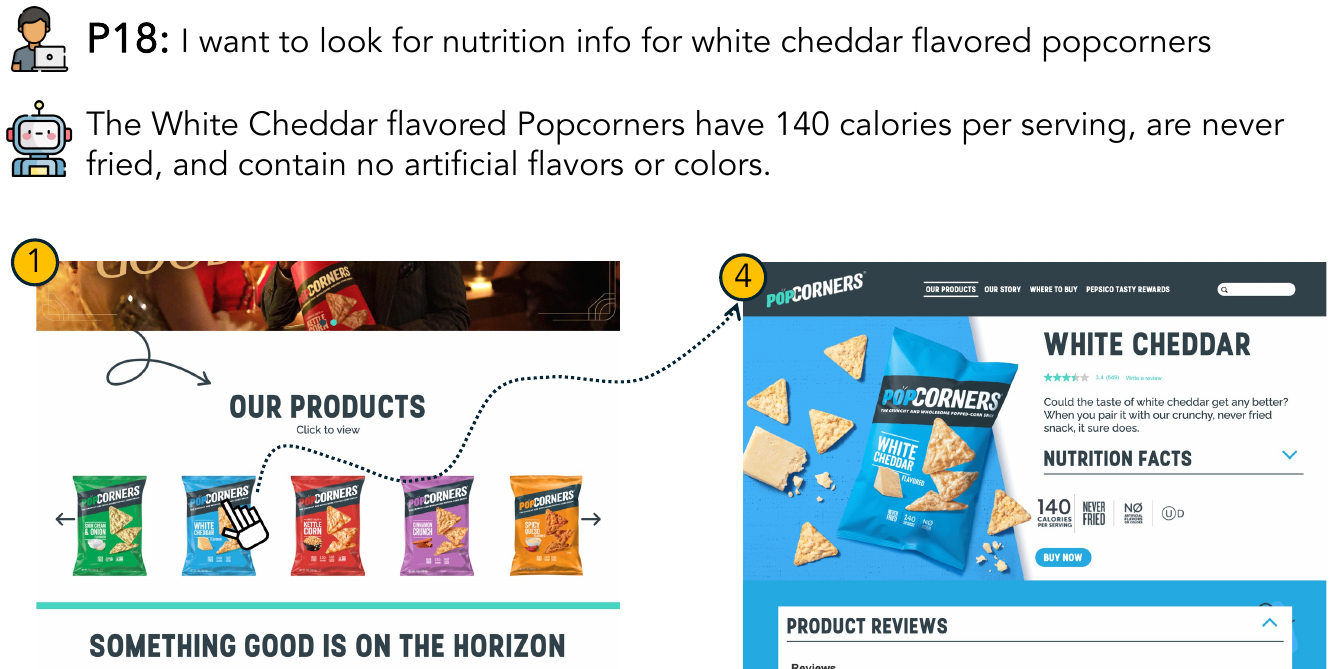}
        \caption{Product and Service Interaction}
        \label{fig:subfig3}
    \end{subfigure}

    \caption{Three example tasks from top three free-form task categories. (all identifiable information has been trimmed for anonymity.)}
    \label{fig:main}
\end{figure}

Each annotator was asked to execute 20 web tasks in collaboration with the LM-based agent. The annotators receive a base payment of $\$0.50$ per task, resulting in a total of $\$10$ (provided as an Amazon gift card, as approved by the IRB of our home institute). Our participants are aged between 20--30 and have varied levels of knowledge about AI agents and varied distribution on daily web tasks.

\begin{table}[t]
\centering
\resizebox{0.8\textwidth}{!}{%
\begin{tabular}{llll}
\toprule
\multicolumn{4}{c}{\textbf{Symbols:} \quad \includegraphics[height=10pt]{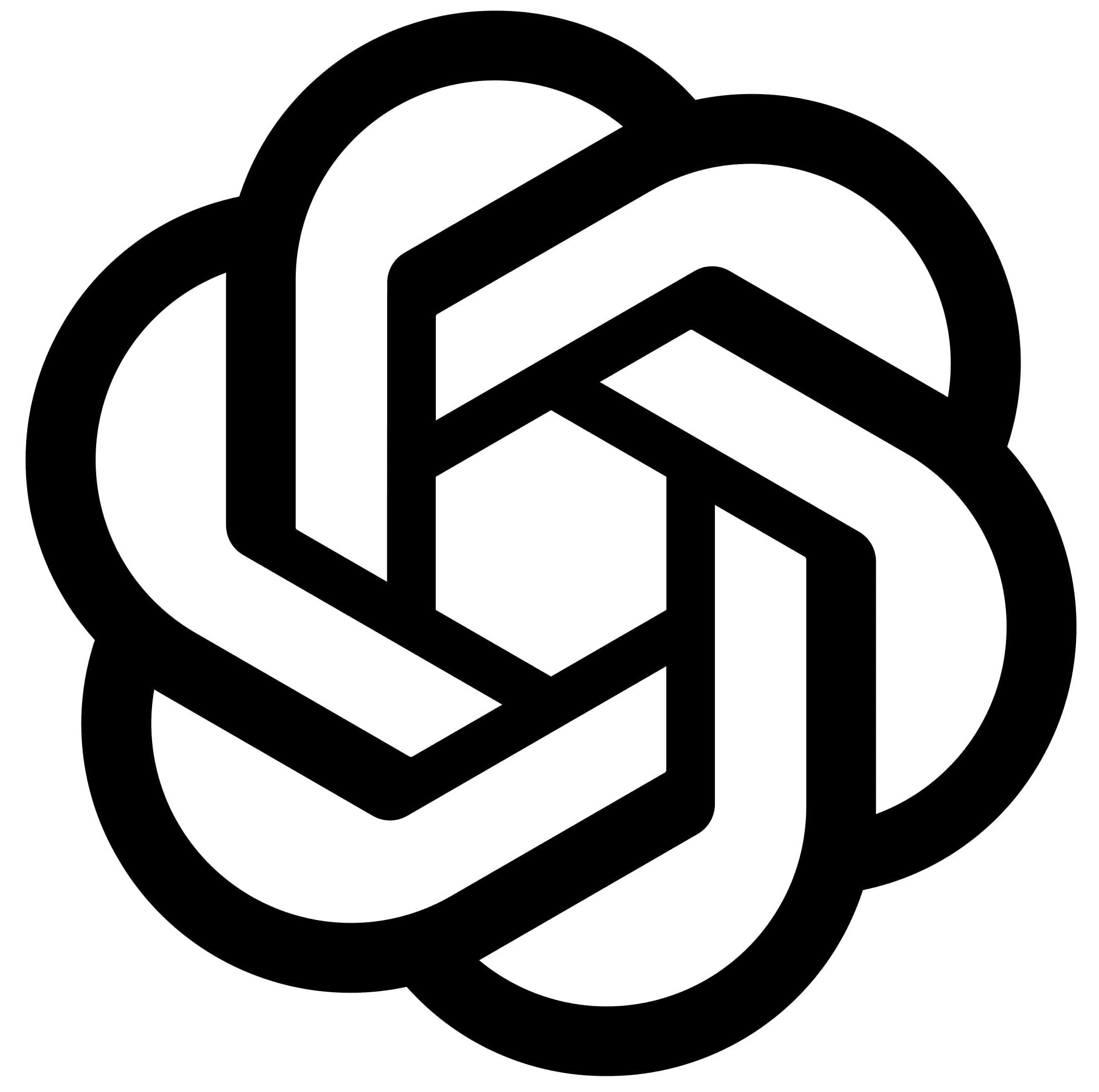} ChatGPT \quad \includegraphics[height=10pt]{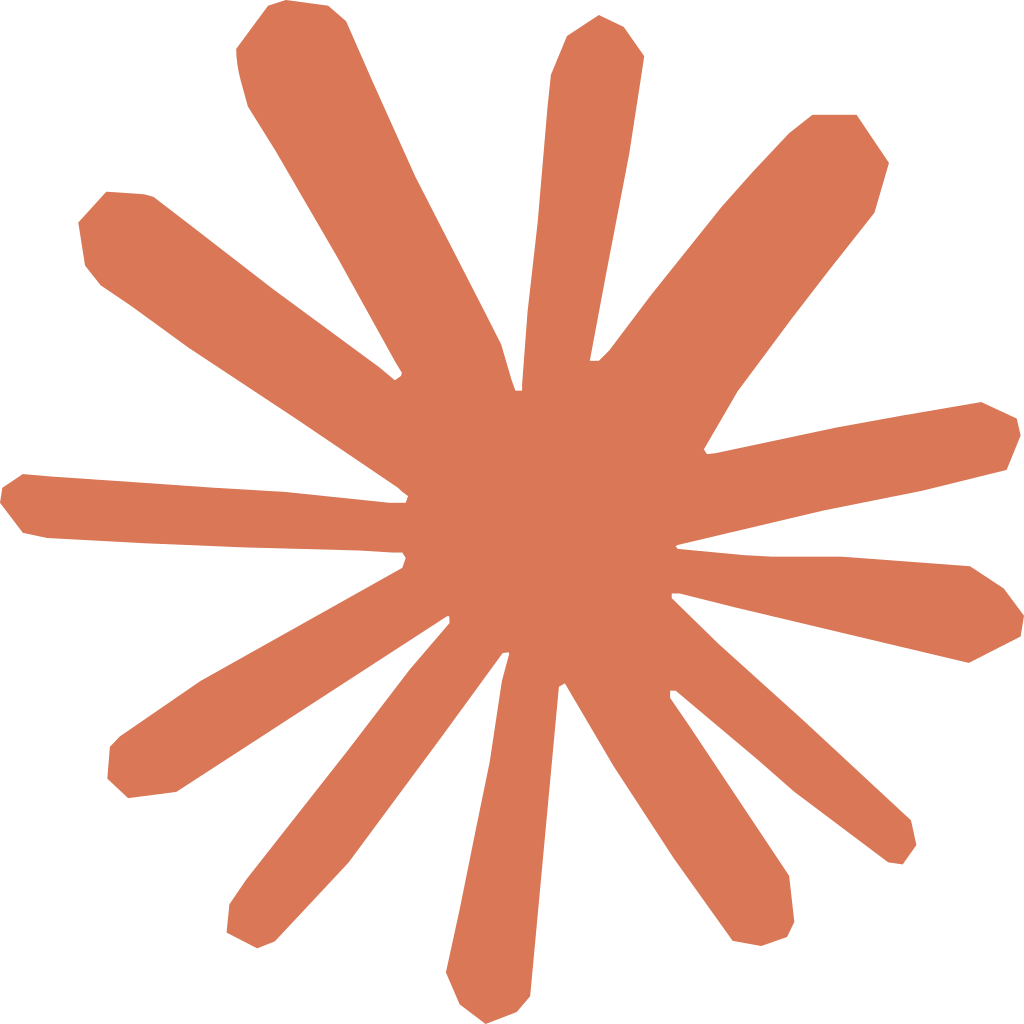} Claude  \quad \includegraphics[height=10pt]{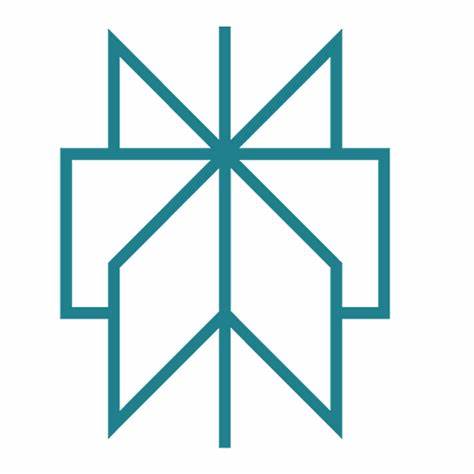} Perplexity \quad \includegraphics[height=10pt]{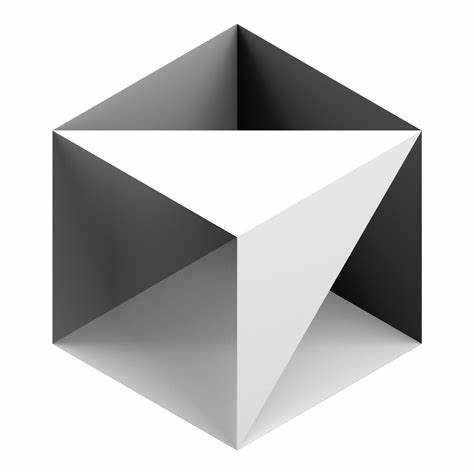} Cursor \quad \includegraphics[height=10pt]{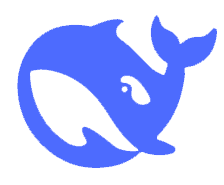} Deepseek \quad \includegraphics[height=10pt]{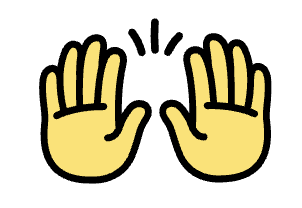} Openhands} \\ \midrule
\textbf{PID} & \textbf{AI usage frequency} & \begin{tabular}[c]{@{}l@{}} \textbf{Familiarity with}\\ \textbf{Agent} (1–7)\end{tabular} & \textbf{Task selection} \\ \midrule
P1 & Few times a day & 3 (\includegraphics[height=9pt]{figures/chatgpt.jpg}, \includegraphics[height=9pt]{figures/perplexity.jpeg}) & \begin{tabular}[c]{@{}l@{}} \tcbox[tabtag=green]{Personalized Interests \& Lifestyle}\tcbox[tabtag=red]{Planning \& Scheduling} \\ \tcbox[tabtag=orange]{Product \& Service Interaction} \end{tabular} \\
P2 & Few times a day & 7 (\includegraphics[height=9pt]{figures/chatgpt.jpg}) & \begin{tabular}[c]{@{}l@{}}\tcbox[tabtag=blue]{Information Access}\end{tabular} \\
P3 & [undisclosed] & [undisclosed] & \begin{tabular}[c]{@{}l@{}}\tcbox[tabtag=blue]{Information Access}\tcbox[tabtag=green]{Personalized Interests \& Lifestyle}\end{tabular} \\
P4 & Few times a day & 2 (\includegraphics[height=9pt]{figures/chatgpt.jpg}, \includegraphics[height=9pt]{figures/perplexity.jpeg}) & \begin{tabular}[c]{@{}l@{}}\tcbox[tabtag=blue]{Information Access}\tcbox[tabtag=green]{Personalized Interests \& Lifestyle}\end{tabular} \\
P5 & Few times a day & 5 (\includegraphics[height=9pt]{figures/chatgpt.jpg}, \includegraphics[height=9pt]{figures/claude.png}) & \begin{tabular}[c]{@{}l@{}}\tcbox[tabtag=blue]{Information Access}\tcbox[tabtag=green]{Personalized Interests \& Lifestyle}\\\tcbox[tabtag=red]{Planning \& Scheduling}\tcbox[tabtag=orange]{Product \& Service Interaction}\end{tabular} \\
P6 & Few times a day & 1 (\includegraphics[height=9pt]{figures/chatgpt.jpg}) & \begin{tabular}[c]{@{}l@{}}\tcbox[tabtag=blue]{Information Access}\tcbox[tabtag=orange]{Product \& Service Interaction}\\\tcbox[tabtag= yellow]{Reasoning \& Meta-Analysis}\end{tabular} \\
P7 & Few times a week & 5 (\includegraphics[height=9pt]{figures/chatgpt.jpg}) & \begin{tabular}[c]{@{}l@{}}\tcbox[tabtag=blue]{Information Access}\tcbox[tabtag=green]{Personalized Interests \& Lifestyle}\end{tabular} \\
P8 & Few times a day & 5 \begin{tabular}[c]{@{}l@{}}(\includegraphics[height=9pt]{figures/chatgpt.jpg}) \end{tabular} & \begin{tabular}[c]{@{}l@{}}\tcbox[tabtag=blue]{Information Access}\tcbox[tabtag=green]{Personalized Interests \& Lifestyle}\end{tabular} \\
P9 & Few times a day & 4 (\includegraphics[height=9pt]{figures/chatgpt.jpg}, \includegraphics[height=9pt]{figures/claude.png}, \includegraphics[height=9pt]{figures/curson.jpeg}) & \begin{tabular}[c]{@{}l@{}}\tcbox[tabtag=blue]{Information Access}\tcbox[tabtag=orange]{Product \& Service Interaction}\end{tabular} \\
P10 & Few times a day & 6 (\includegraphics[height=9pt]{figures/chatgpt.jpg}, \includegraphics[height=9pt]{figures/openhands.png}) & \begin{tabular}[c]{@{}l@{}}\tcbox[tabtag=blue]{Information Access}\end{tabular} \\
P11 & Few times a day & 5 (\includegraphics[height=9pt]{figures/chatgpt.jpg}, \includegraphics[height=9pt]{figures/claude.png}, \includegraphics[height=9pt]{figures/deepseek.png}) & \begin{tabular}[c]{@{}l@{}}\tcbox[tabtag=blue]{Information Access} \tcbox[tabtag=teal]{Content Generation}\\ \tcbox[tabtag=magenta]{Security Testing} \end{tabular} \\
P12 & Few times a day & 6 (\includegraphics[height=9pt]{figures/chatgpt.jpg}) & \begin{tabular}[c]{@{}l@{}}\tcbox[tabtag=teal]{Content Generation}\tcbox[tabtag=blue]{Information Access}\end{tabular} \\
P13 & Few times a day & 5 (\includegraphics[height=9pt]{figures/chatgpt.jpg}) & \begin{tabular}[c]{@{}l@{}}\tcbox[tabtag=blue]{Information Access}\tcbox[tabtag=cyan]{Multi-Agent Collaboration}\\ \tcbox[tabtag= yellow]{Reasoning \& Meta-Analysis}\end{tabular} \\
P14 & Few times a day & 7 (\includegraphics[height=9pt]{figures/chatgpt.jpg}) & \begin{tabular}[c]{@{}l@{}}\tcbox[tabtag=blue]{Information Access}\tcbox[tabtag=cyan]{Multi-Agent Collaboration}\\ \tcbox[tabtag=green]{Personalized Interests \& Lifestyle}\tcbox[tabtag=orange]{Product \& Service Interaction}\end{tabular} \\
P15 & Few times a day & 5 (\includegraphics[height=9pt]{figures/chatgpt.jpg}) & \begin{tabular}[c]{@{}l@{}}\tcbox[tabtag=blue]{Information Access}\tcbox[tabtag=red]{Planning \& Scheduling}\\\tcbox[tabtag=orange]{Product \& Service Interaction}\end{tabular} \\
P16 & Few times a day & 6 (\includegraphics[height=9pt]{figures/chatgpt.jpg}) & \begin{tabular}[c]{@{}l@{}}\tcbox[tabtag=blue]{Information Access}\tcbox[tabtag=cyan]{Multi-Agent Collaboration}\\\tcbox[tabtag=green]{Personalized Interests \& Lifestyle}\\\tcbox[tabtag=red]{Planning \& Scheduling}\tcbox[tabtag=magenta]{Security Testing}\end{tabular} \\
P17 & Few times a day & 2 (\includegraphics[height=9pt]{figures/chatgpt.jpg}) & \begin{tabular}[c]{@{}l@{}}\tcbox[tabtag=teal]{Content Generation}\tcbox[tabtag=blue]{Information Access},\\ \tcbox[tabtag=cyan]{Multi-Agent Collaboration}\end{tabular} \\
P18 & Few times a day & 6 (\includegraphics[height=9pt]{figures/chatgpt.jpg}) & \begin{tabular}[c]{@{}l@{}}\tcbox[tabtag=blue]{Information Access}\end{tabular} \\
P19 & Few times a day & 7 (\includegraphics[height=9pt]{figures/chatgpt.jpg}) & \begin{tabular}[c]{@{}l@{}}\tcbox[tabtag=blue]{Information Access}\tcbox[tabtag=green]{Personalized Interests \& Lifestyle}\end{tabular} \\
P20 & Few times a day & 5 (\includegraphics[height=9pt]{figures/chatgpt.jpg}) & \begin{tabular}[c]{@{}l@{}}\tcbox[tabtag=blue]{Information Access}\end{tabular} \\ \bottomrule
\end{tabular}%
}
\caption{Related backgrounds of participants, including frequency of AI usage, familiarity with AI agents with examples, and the types of tasks executed. Agent familiarity is scored on a 1–7 scale.}
\label{tab:pid}
\end{table}

At the beginning of the study, each participant is onboarded using a walk-through installation video and a detailed description of the study setup. We also offered the participants an opt-in option for installation support, where we helped them to install the agent extension. 11 participants requested the installation support call. On average, users take around 1-1.5 hours to complete the entire annotation. We also provided the participants with a pre-paid API key, so participating in our study did not incur any additional cost to the participants. Participants are given up to 3 chances to execute the given task with the AI agent. Even if a user retries the task multiple times, we keep only one trajectory per task so that it is balanced between each user. In such cases, the participant decides which trajectory they want to submit as their final annotation.

We restricted the free-form tasks to have multiple steps (e.g., not achievable via a single button click or type action). We also encourage users to explore tasks of varied length and complexity.
All our annotators are familiar with the agentic frameworks. \autoref{tab:pid} provides the distribution of each annotator's expertise and choice of tasks. 


We purposefully opt for a self-initiated data annotation paradigm --- i.e., at the end of each task, participants are shown a summary of their annotation, where they can download the data log form if they wish. Such self-initiated data collection ensures \textit{the user has full control over which data they want to share with us and which they do not}.

\subsection{{\sc CowPilot}: Task Annotation Framework used to annotate \cc{}}
\label{sec:3.1:cowpilot-platform}
\begin{figure*}
    \centering
    \includegraphics[width=0.9\linewidth]{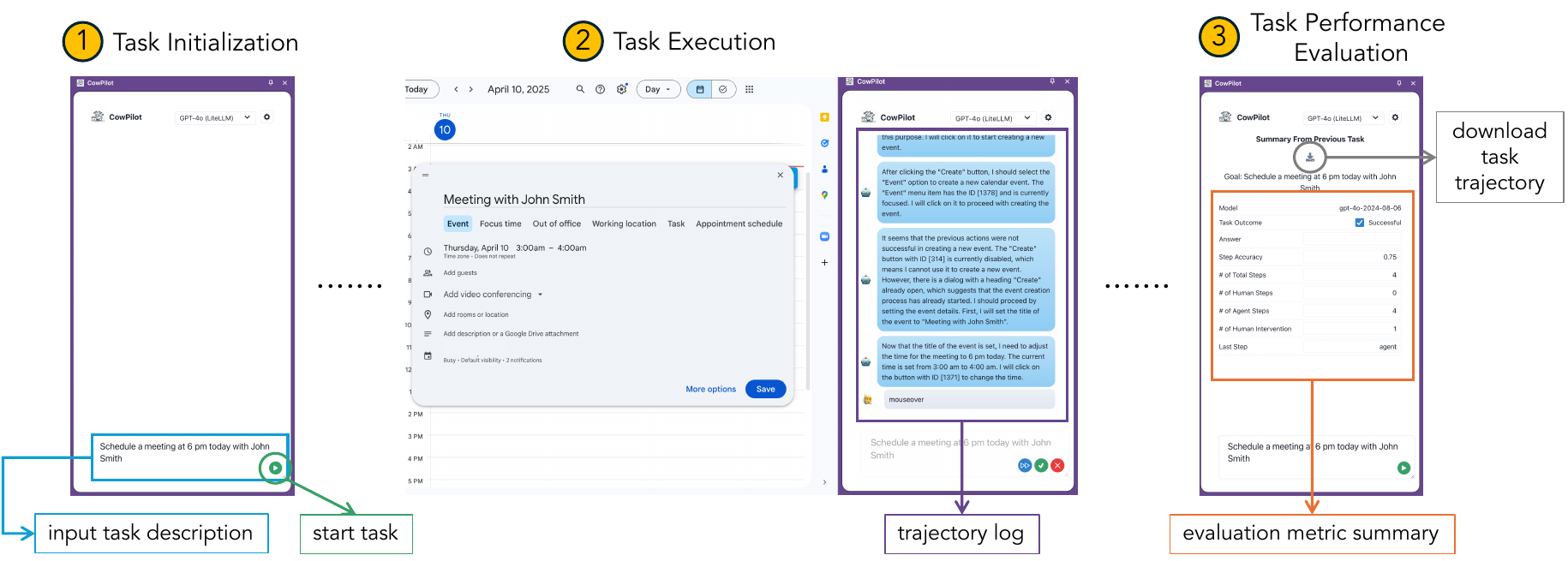}
    \caption{Overview of the collaborative AI agent, {\sc CowPilot} \cite{cowpilot} used in our data collection. 1) Before the task is initiated, the user gives a textual task description as input. 2) During task execution, the system tracks the actions performed by the user and the agent. 3) After the task is executed, the user can download the task log.}
    \label{fig:cowpilot}
\end{figure*}

We select \cp \citep{cowpilot}, an open-sourced Chrome extension for collaborative web navigation between a human and an LLM-powered agent (\autoref{fig:cowpilot}), to annotate our data. We chose \cp since it can be downloaded as a Chrome extension and easily integrated into the users' current browsing workflow. The AI agent is instantiated using LLM, which is capable of extracting the web HTML and generating a step-by-step task execution plan on real-world websites.

The system offers a \textit{suggest-then-execute} workflow where the AI agent proposes UI actions (e.g., clicking buttons, filling text boxes, going to a specific URL) that are visually highlighted for user approval. Users can allow the agent to proceed, pause its execution, or intervene in the agent and take over control. Users can intervene at arbitrary times and for an unlimited number of times if they wish. The entire interaction between the user and the agent, as well as the web environment information, can be logged in detail, capturing both human and agent actions for later analysis --- making the system an ideal candidate to collect \cc.

While CowPilot supports a wide range of LLM backends, we used \texttt{GPT-4o} in our study for consistency across annotators. Participants primarily used CowPilot in the \textit{copilot mode}, as both agent automation and human intervention are important.

\subsection{Analysis}
\subsubsection{Users Collaborate on Versatile Tasks with Agents}
\label{sec:4.3:task-dist}

The participants picked a wide range of free-form tasks in {\sc CowCorpus}. To understand which kinds of tasks a collaborative agent is most impactful with, we categorize the free-form tasks into the 9 categories in \autoref{tab:example_free_form}, and annotate common tasks performed by individual users in \autoref{tab:pid}.

\begin{itemize}
\itemindent=-13pt
    \item \tcbox[on line, tabtag=blue]{Information Access}: Users are looking for specific information, such as facts, news, academic papers, and definitions.
    \item \tcbox[on line, tabtag=green]{Personalized Interests \& Lifestyle}: Engagement with entertainment or lifestyle content.
    \item \tcbox[on line, tabtag=orange]{Product \& Service Interaction}: Users shop for specific products, compare prices, or book services like flights or rentals. 
    \item \tcbox[on line, tabtag=teal]{Content Generation}: Users want to compose written content for communication such as social media posts and emails.
    \item \tcbox[on line, tabtag=cyan]{Multi-Agent Collaboration}: Tasks where users coordinate actions across multiple AI tools — such as delegating subtasks to ChatGPT or combining system outputs to accomplish a goal.
    \item \tcbox[on line, tabtag=magenta]{Security Testing}: Tasks designed to probe the agent’s robustness or ethical boundaries — including prompt injections, adversarial inputs, or potentially inappropriate or harmful requests.
    \item \tcbox[on line, tabtag=red]{Planning \& Scheduling}: Tasks centered on organizing time-bound activities — such as scheduling meetings or creating events.
    \item \tcbox[on line, tabtag= yellow]{Reasoning \& Meta-Analysis}: Tasks that require analytical thinking or synthesis — such as comparing models, summarizing research contributions, or evaluating options beyond basic retrieval.
    \item \tcbox[on line, tabtag= gray]{Misc.} Tasks that do not clearly align with any of the above categories. These may include underspecified commands or uncommon task types.
\end{itemize}

Three authors followed an open-coded approach to develop the final categories. Throughout the process, we followed standard practices from past works \cite{deng2024mind2web} and incorporated security \cite{liao2025eiaenvironmentalinjectionattack} and multi-agent collaborative aspects \cite{han2024llmmultiagentsystemschallenges} of AI agents. 


\subsubsection{Current Agent Bottleneck: Time Demand}
\label{sec:4.4:time-log}

While collaboration yielded benefits in terms of control and success, it also introduced additional time demands. As shown in \autoref{tab:eval_result}, in \cp{}, agent execution took on average $93.1$ seconds for standard tasks and $71.7$ seconds for free-form tasks. In contrast, human intervention time was relatively short—$23.9$ seconds and $13.8$ seconds, respectively 

On the other hand, from post-annotation ratings, participants gave a neutral rating of $4.05$ when asked if they completed the task faster than they would have without \cp{} (\autoref{fig:rating}, last row), suggesting uncertainty about whether relying on the agent actually saved time for users, compared to executing the tasks themselves.

\begin{figure}[h]
    \centering
    \includegraphics[width=0.9\linewidth]{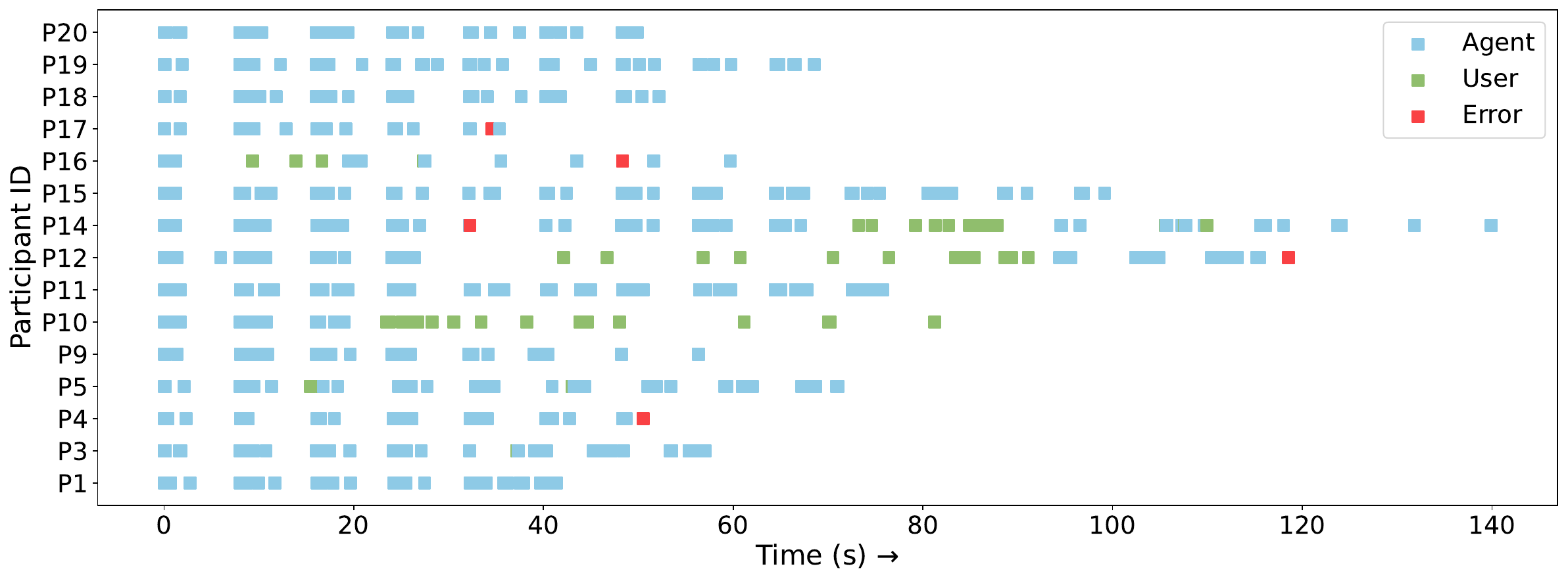}
    \caption{Time log across participants for the same task}
    \label{fig:time_log}
\end{figure}

This agent-heavy time distribution reflects current limitations of LLM-based agents. Each step in the task sequence involves non-trivial latency, and as shown in the time log (\autoref{fig:time_log}), agents proceed at a constant pace unless interrupted. These delays accumulate especially in longer-horizon tasks. Moreover, users need to continuously monitor the agent and be ready to intervene, requiring sufficient observation windows to inspect each step, which further slows down the overall process. 

On the other hand, with \cpnew{}, we see a significant increase in time requirement (an average score of $5.25$). Users were more satisfied with the updated interaction module where the agent only intervened as needed rather than the continiously needing to monitor it. They also gave a higher rating of $5.75$ that \cpnew{} avoided interrupting them unnecessarily.


These findings highlight a key limitation of current agents: the \textbf{inability to proactively request help}. Future collaborative agents could incorporate uncertainty estimation mechanisms to identify decision points where user input is most valuable, rather than maintaining a fixed execution pace throughout the task.

\subsubsection{Full Benchmark Table}
\label{sec:4.5:benchmark-table}
We present the comprehensive evaluation results across all metrics in \autoref{tab:all-data-results}, including Precision and Recall which were omitted from the main text for brevity. Our fine-tuned model (Gemma 27B) demonstrate the most balanced performance, maintaining high non-intervention accuracy while significantly improving intervention recall compared to their base counterparts and closed source models.

\begin{table}[H]
\resizebox{\columnwidth}{!}{%
\begin{tabular}{lcccccccccc}
\hline
 & \multirow{2}{*}{\textbf{\begin{tabular}[c]{@{}c@{}}Step \\ Acc\end{tabular}}} & \multicolumn{2}{c}{\textbf{Precision}} & \multicolumn{1}{l}{} & \multicolumn{2}{c}{\textbf{Recall}} & \multicolumn{1}{l}{} & \multicolumn{2}{c}{\textbf{F1 Score}} & \multirow{2}{*}{\textbf{PTS}} \\ \cline{3-4} \cline{6-7} \cline{9-10}
 &  & \begin{tabular}[c]{@{}c@{}}Interv\end{tabular} & \begin{tabular}[c]{@{}c@{}}Non-Interv\end{tabular} &  & \begin{tabular}[c]{@{}c@{}}Interv\end{tabular} & \begin{tabular}[c]{@{}c@{}}Non-Interv\end{tabular} &  & \begin{tabular}[c]{@{}c@{}}Interv\end{tabular} & \begin{tabular}[c]{@{}c@{}}Non-Interv\end{tabular} &  \\ \hline
 \multicolumn{1}{c}{\textit{Baselines}} &  &  &  &  &  &  &  &  &  &  \\ \cline{1-1}
Always Interv & 0.147 & 0.147 & 0.000 &  & 1.000 & 0.000 &  & 0.257 & 0.000 & 0.151 \\
Always No Interv & 0.853 & 0.000 & 0.853 &  & 0.000 & 1.000 &  & 0.000 & 0.920 & 0.000 \\
\hline
\multicolumn{1}{c}{\textit{Closed Source Models}} &  &  &  &  &  &  &  &  &  &  \\ \cline{1-1}
\texttt{Claude 4 Sonnet} &  &  &  &  &  &  &  &  &  &  \\
0 shot (w/o reasoning) & 0.681 & 0.179 & 0.864 &  & 0.324 & 0.743 &  & 0.231 & 0.799 & \textbf{0.293} \\
0 shot (w/ reasoning) & 0.697 & 0.169 & 0.859 &  & 0.270 & 0.771 &  & 0.208 & 0.813 & 0.164 \\
2 shot (w/o reasoning) & \textbf{0.749} & 0.158 & 0.854 &  & 0.162 & \textbf{0.850} &  & 0.160 & \textbf{0.852} & 0.149 \\
2 shot (w/ reasoning) & 0.721 & 0.149 & 0.853 &  & 0.189 & 0.813 &  & 0.167 & 0.833 & 0.158 \\

\texttt{GPT-4o} &  &  &  &  &  &  &  &  &  &  \\
0 shot (w/o reasoning) & 0.741 & 0.182 & 0.860 &  & 0.216 & 0.832 &  & 0.198 & 0.846 & 0.147 \\
2 shot (w/o reasoning) & 0.661 & 0.147 & 0.852 &  & 0.270 & 0.729 &  & 0.190 & 0.786 & 0.206 \\

\texttt{Gemini 2.5 Pro} &  &  &  &  &  &  &  &  &  &  \\
0 shot (w/o reasoning) & 0.681 & 0.213 & 0.881 &  & 0.432 & 0.724 &  & 0.286 & 0.795 & 0.262 \\
0 shot (w/ reasoning) & 0.689 & 0.211 & 0.878 &  & 0.405 & 0.738 &  & 0.278 & 0.802 & 0.237 \\
2 shot (w/o reasoning) & 0.586 & 0.181 & 0.877 &  & 0.514 & 0.598 &  & 0.268 & 0.711 & 0.287 \\
2 shot (w/ reasoning) & 0.641 & \textbf{0.221} & \textbf{0.897} &  & \textbf{0.568} & 0.654 &  & \textbf{0.318} & 0.757 & 0.243 \\

\hline
\multicolumn{1}{c}{\textit{Open Source Models}} &  &  &  &  &  &  &  &  &  &  \\ \cline{1-1}
\texttt{Gemma 27B (base)} &  &  &  &  &  &  &  &  &  &  \\
0 shot (w/o reasoning) & 0.239 & 0.154 & \textbf{0.897} &  & \textbf{0.919} & 0.121 &  & 0.264 & 0.214 & 0.187 \\

\texttt{Llava 8B (base)} &  &  &  &  &  &  &  &  &  &  \\
0 shot (w/o reasoning) & 0.183 & 0.000 & 0.852 &  & 0.000 & 0.215 &  & 0.000 & 0.343 & 0.017 \\

\hline
\multicolumn{1}{c}{\textit{Our Models}} &  &  &  &  &  &  &  &  &  &  \\ \cline{1-1}

\texttt{Gemma 27B (finetuned)} &  &  &  &  &  &  &  &  &  &  \\
0 shot (w/o reasoning) & 0.853 & \textbf{0.500} & 0.877 &  & 0.216 & \textbf{0.963} &  & \textbf{0.302} & \textbf{0.918} & \textbf{0.303} \\

\texttt{Llava 8B (finetuned)} &  &  &  &  &  &  &  &  &  &  \\
0 shot (w/o reasoning) & \textbf{0.817} & 0.471 & 0.876 &  & 0.216 & 0.921 &  & 0.296 & 0.897 & 0.201 \\

\hline
\end{tabular}%
}
\caption{Step Accuracy, Precision, Recall, F1-score and PTS score for Human Intervention Prediction Task on All Data. }
\label{tab:all-data-results}
\end{table}

\section{Ablation on Modeling Human Intervention}

\subsection{Ablation: Few Shot Example Count}
We investigate whether providing in-context examples (2-shot) improves intervention timing. As shown in \autoref{tab:ablation-shots}, the impact of few-shot prompting is inconsistent across models. While 2-shot prompts slightly improve PTS for GPT-4o and Gemini, they significantly degrade Claude's intervention timing. This suggests that for certain models, few-shot examples might introduce bias or over-constrain the model's decision boundary, making zero-shot the more robust setting for this specific task.
\begin{table}[h]
\centering
\resizebox{0.6\columnwidth}{!}{%
\begin{tabular}{lcccccc}
\toprule
\multirow{2}{*}{\textbf{Model}} & \multirow{2}{*}{\textbf{Shot}} & \textbf{Step} & \multicolumn{2}{c}{\textbf{F1 Score}} & \multirow{2}{*}{\textbf{PTS}} \\ \cmidrule(lr){4-5}
 &  & \textbf{Acc} & \textbf{Interv} & \textbf{Non-Interv} &  \\ \midrule
\multirow{2}{*}{\texttt{Claude 4 Sonnet}} & 0-shot & 0.681 & \textbf{0.231} & 0.799 & \textbf{0.293} \\
 & 2-shot & \textbf{0.749} & 0.160 & \textbf{0.852} & 0.149 \\ \midrule
\multirow{2}{*}{GPT-4o} & 0-shot & \textbf{0.741} & \textbf{0.198} & \textbf{0.846} & 0.147 \\
 & 2-shot & 0.661 & 0.190 & 0.786 & \textbf{0.206} \\ \midrule
\multirow{2}{*}{\texttt{Gemini 2.5 Pro}} & 0-shot & \textbf{0.681} & \textbf{0.286} & \textbf{0.795} & 0.262 \\
 & 2-shot & 0.586 & 0.268 & 0.711 & \textbf{0.287} \\ \bottomrule
\end{tabular}%
}
\caption{Ablation on Few-Shot Setting (without reasoning)}
\label{tab:ablation-shots}
\end{table}

\subsection{Ablation: Models with Reasoning vs No Reasoning}
The result from  \autoref{tab:ablation-reasoning} reveals a counter-intuitive finding: explicit reasoning tends to lower the Perfect Timing Score (PTS) across models. While reasoning slightly improves Step Accuracy by reducing false positives, it also strengthens the models' bias toward staying silent. This suggests that human intervention is often a reactive, intuitive decision, and forcing a model to articulate a logical justification result in hesitant agents that intervene too late.
\begin{table}[h]
\centering
\resizebox{0.6\columnwidth}{!}{%
\begin{tabular}{lcccccc}
\toprule
\multirow{2}{*}{\textbf{Model}} & \multirow{2}{*}{\textbf{Reasoning}} & \textbf{Step} & \multicolumn{2}{c}{\textbf{F1 Score}} & \multirow{2}{*}{\textbf{PTS}} \\ \cmidrule(lr){4-5}
 &  & \textbf{Acc} & \textbf{Interv} & \textbf{Non-Interv} &  \\ \midrule
\multirow{2}{*}{\texttt{Claude 4 Sonnet}} & No & 0.681 & \textbf{0.231} & 0.799 & \textbf{0.293} \\
 & Yes & \textbf{0.697} & 0.208 & \textbf{0.813} & 0.164 \\ \midrule
\multirow{2}{*}{\texttt{Gemini 2.5 Pro}} & No & 0.681 & \textbf{0.286} & 0.795 & \textbf{0.262} \\
 & Yes & \textbf{0.689} & 0.278 & \textbf{0.802} & 0.237 \\ \bottomrule
\end{tabular}%
}
\caption{Ablation on Reasoning Setting (0 shot)}
\label{tab:ablation-reasoning}
\end{table}

\subsection{Ablation: Impact of Human Action History}
\begin{wraptable}[8]{r}{0.5\textwidth}
\centering
\resizebox{0.4\textwidth}{!}{%
\begin{tabular}{ccc}
\hline
\textbf{\begin{tabular}[c]{@{}c@{}}Includes Human \\ Action History\end{tabular}} & 
\textbf{\begin{tabular}[c]{@{}c@{}}Step \\ Accuracy\end{tabular}} & 
\textbf{Macro F1} \\ \hline
\cmark &  \textbf{0.8136} & 0.4486 \\ \hline 
\xmark & 0.7627 & 0.4327 \\  \hline
\end{tabular}%
}
\caption{Ablation on the inclusion of human actions (\texttt{Claude} 2 shot w/o reasoning).}
\label{tab:ablation-2}
\vspace{20pt}
\end{wraptable}
To understand the importance of temporal context in collaborative tasks, we conduct an ablation study by removing the history of human actions from the agent's input.
As shown in \autoref{tab:ablation-2}, explicitly including the human action history improves model performance, increasing Step Accuracy from 76.27\% to 81.36\%. This gain suggests that the agent's decision-making process is not solely dependent on the current state observation but is sensitive to the trajectory of past interactions.

\subsection{Ablation: Impact of Input Format on Performance}
\begin{wraptable}{r}{0.5\textwidth}
\vspace{-10pt} 
\centering
\resizebox{0.5\textwidth}{!}{%
\begin{tabular}{cccc}
\hline
\multicolumn{1}{c}{\textbf{\begin{tabular}[c]{@{}c@{}}Includes\\ Screenshot?\end{tabular}}} & \multicolumn{1}{c}{\textbf{\begin{tabular}[c]{@{}c@{}}Includes\\ AXTree?\end{tabular}}} & \textbf{\begin{tabular}[c]{@{}c@{}}Step \\ Accuracy\end{tabular}} & \textbf{Macro F1} \\ \hline
\xmark & \cmark & 0.697 &  0.208 \\ \hline
\cmark & \xmark & 0.729 &  0.171 \\ \hline
\cmark & \cmark & 0.749 & 0.160 \\ \hline
\end{tabular}%
}
\caption{Ablation on input format}
\label{tab:ablation-3}
\vspace{-10pt}
\end{wraptable}
We further investigate the contribution of different observation modalities by evaluating the agent's performance when restricted to either visual inputs (Screenshots) or structural text inputs (AXTree). 

As shown in \autoref{tab:ablation-3}, the Screenshot-only and AXTree-only approaches yield Step Accuracies of 72.9\% and 69.7\%, respectively. Both are lower than the multimodal baseline (74.9\%), demonstrating the benefit of combining visual and structural information.

\subsection{Intervention Prediction Remains Robust Under Time Offsets}
\label{subsec:PTS}

Across a wide range of $\alpha$ values, PTS provides a consistent and stable metric for comparing model performance. In the zero-shot setting, we observe that increasing $\alpha$ leads to a monotonic decrease in PTS scores across all baseline models, including those equipped with explicit reasoning capabilities, while preserving their relative ranking, as confirmed by a high Kendall’s W. This demonstrates that PTS maintains consistent model ranking across variations of $\alpha$ and reliably reflects the underlying quality of a model’s intervention prediction.

\begin{figure*}[h]
    \centering
    \begin{minipage}{0.47\textwidth}
        \centering
        \includegraphics[width=\linewidth]{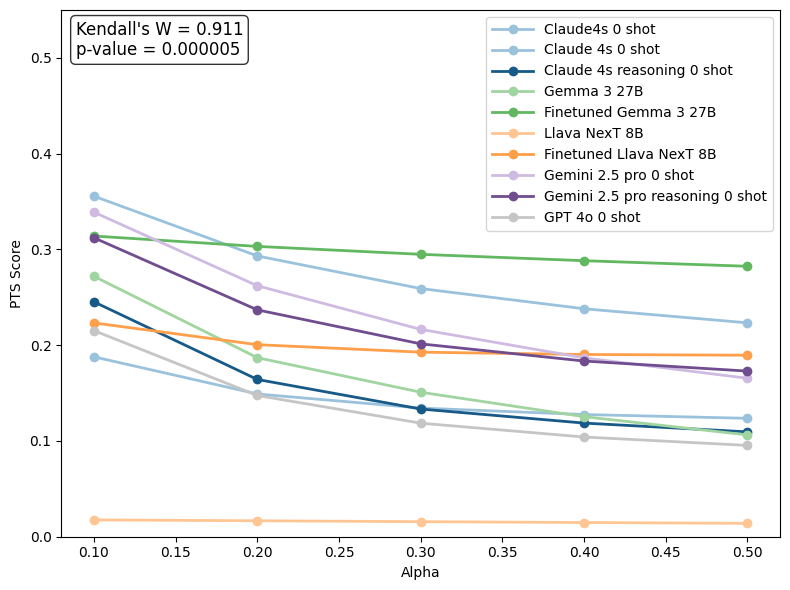}
        \caption{Zero-shot models maintain consistent PTS rankings across $\alpha$. We sweep $\alpha$ over a fixed grid while holding all inputs constant and recompute PTS for each model under zero-shot setting. Kendall's W significant test reveals that PTS preserves relative ordering under different temporal penalties. }
        \label{fig:ablation-zero-shot}
    \end{minipage}
    \hfill
    \begin{minipage}{0.47\textwidth}
        \centering
        \includegraphics[width=\linewidth]{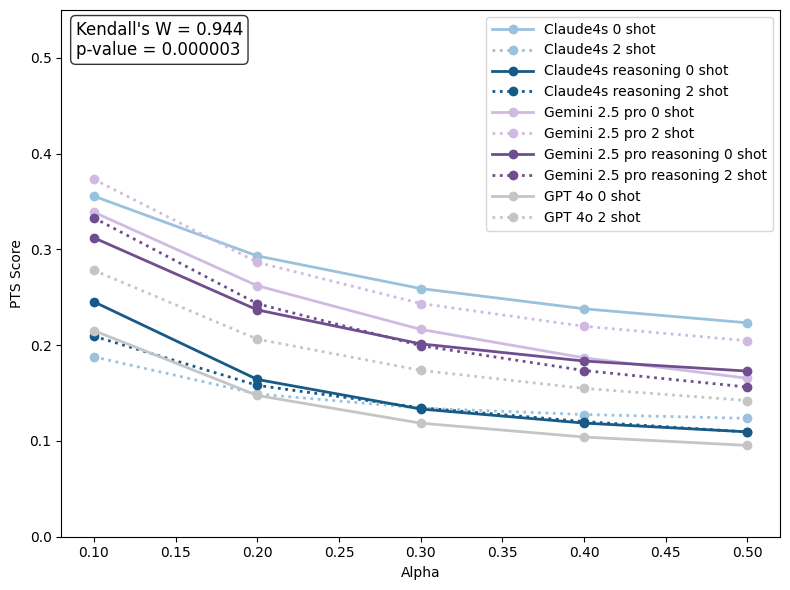}
        \caption{Closed-source models maintain consistent PTS rankings across $\alpha$. We sweep $\alpha$ over a fixed grid while holding all inputs constant and recompute PTS for each closed-source model. Kendall’s W significant test reveals that PTS preserves relative ordering under different temporal penalties. }
        \label{fig:ablation-close}
    \end{minipage}
\end{figure*}


Notably, the PTS curves of fine-tuned models show much stable to changes in $\alpha$. Since $\alpha$ controls how strongly early or mistimed predictions are penalized, a model whose intervention timing is already close to the ground truth will accumulate only small penalties regardless of the exact $\alpha$ value. The relative flatness of the fine-tuned curves therefore indicates that these models consistently make temporally accurate predictions with fewer premature or missed calls.


\end{document}